\documentclass[conference]{IEEEtran}
\IEEEoverridecommandlockouts
\renewcommand\IEEEkeywordsname{Index Terms}
\usepackage{tabularx}
\usepackage{fancyhdr}
\usepackage{cite}
\usepackage{amsmath,amssymb,amsfonts}
\usepackage{algorithmic}
\usepackage{graphicx}
\usepackage{textcomp}
\usepackage{balance}
\usepackage{comment}
\usepackage{multicol}
\usepackage{multirow}
\usepackage{url}

\usepackage{xcolor}
\def\BibTeX{{\rm B\kern-.05em{\sc i\kern-.025em b}\kern-.08em
    T\kern-.1667em\lower.7ex\hbox{E}\kern-.125emX}}
\begin{document}

\title{Efficient Golf Ball Detection and Tracking Based on Convolutional Neural Networks and Kalman Filter\\
}

\author{\IEEEauthorblockN{Tianxiao Zhang$^{{a}}$, Xiaohan Zhang$^{a}$, Yiju Yang$^a$, Zongbo Wang$^b
$, Guanghui Wang$^{c\ddag}$}
\IEEEauthorblockA{$^a$ \textit{Department of Electrical Engineering and Computer Science, University of Kansas, Lawrence, KS 66045 USA}\\
$^b$ \textit{Ainstein Inc., Lawrence, Kansas, 66047, USA}\\
$^c$ \textit{Department of Computer Science, Ryerson University, Toronto, ON, Canada M5B 2K3}\\
$^\ddag$ Corresponding author. wangcs@ryerson.ca
}}
\maketitle
\begin{abstract}

This paper focuses on the problem of online golf ball detection and tracking from image sequences. An efficient real-time approach is proposed by exploiting convolutional neural networks (CNN) based object detection and a Kalman filter based prediction. Five classical deep learning-based object detection networks are implemented and evaluated for ball detection, including YOLO v3 and its tiny version, YOLO v4, Faster R-CNN, SSD, and RefineDet. The detection is performed on small image patches instead of the entire image to increase the performance of small ball detection. At the tracking stage, a discrete Kalman filter is employed to predict the location of the ball and a small image patch is cropped based on the prediction. Then, the object detector is utilized to refine the location of the ball and update the parameters of Kalman filter. In order to train the detection models and test the tracking algorithm, a collection of golf ball dataset is created and annotated. Extensive comparative experiments are performed to demonstrate the effectiveness and superior tracking performance of the proposed scheme.

\end{abstract}

\section{Introduction}
\label{sec:introduction}
Golf is a popular sport that has attracted a huge number of participants and audiences. However, watching the dim, tiny, fast-moving golf balls flying around is not such a fun experience. Practicing golf in the golf driving range causes the problem to visualize the trajectory of the golf ball. In certain circumstances, the golf ball quickly flies beyond the visual sight immediately after being hit. That is why golf is one of the sports that frequently utilize video analysis during the training session. However, detecting the fast-moving tiny golf ball is very challenging and problematic, since the golf ball can fly extremely fast and quickly disappear from the camera field of view. As soon as the golf ball flies farther from the camera, it becomes smaller and smaller in the image, which increases the difficulty in detection. Another huge challenge is caused by motion blur. As we know that, if the camera's frame rate is not high enough, a significant blur will appear in the image, making the ball hard to detect.  

Most of the proposed method could be classified as sensor integration method\ cite{article}, traditional object tracking method \cite{7279260}, and traditional computer vision approach \cite{Kim2011AutomaticHF}\cite{CV}. 
Umek \textit{et al}. \cite{article} proposed a sensor integrated golf equipment. Without changing the functionality of a golf club, two orthogonal affixed train gage (SG) sensors, a 3-axis gyroscope, and an accelerometer were used to monitor the state and actions of the golf swing. Different types of golf swings and movement in early phases of swing can be detected with strain gage sensors. By collecting the returned data, the biofeedback application could be used to help golf beginners to learn repetitive swing. For the traditional object tracking approach, Lyu \textit{et al.} \cite{7279260} proposed a real-time high speed moving ball shape object tracking algorithm using the fusion of multiple features. In the initial step, frame difference is obtained by subtracting the two consecutive frames. The image obtained is converted to a binary image and an array of candidate objects is collected based on all the candidate contours. Then, the moments of all the contours are calculated. False candidates are eliminated by applying a multi-feature extraction method. Finally, the ROI is refined by the contour obtained in the previous step. 

Using the traditional computer vision approach, Woodward \textit{et al}. \cite{CV} proposed a low-cost golf ball tracking method. The images are taken by two calibrated stereo cameras and reconstructed in 3-D space. To realize VR golf, the authors only consider recognizing the golf ball that is centimeter-wise from the camera. In the system, there are two dots, one blue and one yellow, attached to the golf club, and the golf ball is pink. The system requires an unchanged background. By subtracting the consecutive frame by the background, the golf ball could be tracked. While in reality, golf balls are a hundred meters away from the camera. The 3D reconstruction process is computationally intensive, which could not be implemented in real-time. The authors labeled the golf ball and the club with different colors, by recognizing the color feature, the club and golf ball could be detected. While, overall, none of these are efficient for real-time golf ball detection.  Kim \textit{et al}. \cite{Kim2011AutomaticHF} proposed a two-stage method for ball detection from images taken by the multi-exposure camera. They first estimate the region of the ball based on a threshold calculated using Otsu's method. Then, segment the ball through a labeling process. 

Classical visual object tracking algorithms all fail in tracking the golf ball since the ball is too small in the image. In our experiments, we found that feeding high-quality images directly into the detection models led to significant computational load and terrible results. Inspired by \cite{7279260}, we propose an accurate and real-time golf ball tracking approach based on the object detection and Kalman filter \cite{Wu_Vision2017}. Instead of using the entire image, we make detection from small cropped image patches to increase the detection accuracy without sacrificing speed. Extensive results demonstrate the effectiveness of the proposed approach. 

The main contributions of the paper include
\begin{enumerate}
    \item The paper proposes an effective and efficient approach for real-time detection and tracking of a small golf ball based on object detection and Kalman filter.
    \item Six CNN-based detection models have been implemented and compared in terms of detection precision and speed.
    \item A dataset with labeled golf ball images and sequences is generated to train and test the proposed approach.
\end{enumerate}
A short version of this paper has been accepted for publication at SMC 2020 conference \cite{9283312}. The manuscript has made substantial extensions of the conference version by adding more models, datasets, and experiments. The proposed method and the labeled dataset will be available on the author's GitHub site: \url{https://github.com/rucv/golf_ball}.

\section{Related Work}
Artificial intelligence (AI) has significantly changed our lives in recent years. In the field of sports, AI technology is opening up a new path. NBA has established a data computing system that can mine data in games and model data through machine learning. AI is also starting to play the role of auxiliary training. Microsoft has developed a sports performance platform, a set of data management system that analyzes the training and performance of athletes. Computer vision techniques can provide athletes with sports data analysis, help the coach to target weak areas, and improve sports performance, such as pose estimation, object detection, and object tracking. Golf is one of the sports that frequently utilize video analysis and could be benefited significantly with the help of computer vision techniques. Pose estimation could assist to improve the action of the swing. Object tracking could help to track and plot the curve of the golf ball for further analysis. 

To better assist golf players during training, Umek \textit{et al}. \cite{article} proposed a method based on sensor fusion to record the swing data. Wang \textit{et al}. \cite{8474387} proposed a high-speed stereo vision system under indoor lighting conditions using circle detection \cite{8296246} to detect the golf ball and utilize the dynamic ROI for golf ball tracking. Lyu \textit{et al}. \cite{7279260} proposed a golf ball tracking method based on multiple features. In real circumstances, the illumination variations, weather conditions, and background variations will complicate the detection and tracking of the golf ball. Inspired by the progress of the state-of-the-art object detectors, which have demonstrated great capability in daily lives, such as automatic driving, target tracking using surveillance cameras. We propose to solve this problem by using deep learning models.

Object detection and tracking are two classical computer vision problems. With the fast development of deep learning, convolutional neural network (CNN) based object detection and tracking approaches have drawn more attention over the classical approaches due to their unprecedented performance. Based on the tremendous development of computing power, especially GPU, which makes it possible to detect and track an object in real-time \cite{li2019siamrpn++}. Object tracking is the process of locating a moving object or (multiple objects) over time in consecutive video frames. Object tracking is widely applied in human-computer collaboration, traffic monitoring, and surveillance system. Classical tracking methods \cite{10.1109/CVPR.2014.143}\cite{7173015}\cite{sui2016tracking} require manual initialization with the ground truth in the first frame. In recent years, the tracking-by-detection methods have drawn more attention for their real-time applications with the fast development of GPUs and TPUs. Correlation filter-based trackers have also attracted a lot of attention due to their high-speed performance \cite{henriques2014high}\cite{sui2018joint}\cite{sui2019exploiting}. While when it comes to this specific problem, with such a small and extremely fast-moving object, they all fail and would return useless information once failed during tracking. 

Object detection focuses on detecting semantic instances for certain classes in videos or images. The fundamental purpose of object detection is to simultaneously localize and classify the objects shown in the images or videos. Object detection plays a crucial role in lots of practical applications, such as face recognition, pose estimation, medical diagnostics, etc. Object detection has a variety of applications, however, it also comes with challenges and problems. Some commonly seen challenges and problems are variations of object sizes, occlusions, viewpoints, and light conditions. A number of papers attempt to solve those problems, especially how to detect small objects in images or videos. The methods for object detection generally fall into either classical machine learning approaches \cite{DBLP:conf/cvpr/DalalT05} or deep learning approaches \cite{bochkovskiy2020yolov4}\cite{li20202}\cite{lin2017feature}\cite{lin2017focal}\cite{ma2020mdfn}\cite{DBLP:journals/corr/RedmonF16}\cite{Redmon2018YOLOv3AI}\cite{zhang2018single}. In classical machine learning based approaches, the features are predefined, while deep learning based methods are able to perform end-to-end training without specifically defined features. In golf ball detection, the machine learning based approach does not show good performance, although the golf ball has a distinctive shape and color. In practical circumstances, there are many distracting objects in the scene, which could cause failure in target object detection. In addition, the background in real circumstances is not stationary, making it difficult to apply the background subtraction approach. 

Deep learning based object detection models are usually categorized as one-stage methods and two-stage methods. For two-stage methods, the model proposes a set of candidates by selective search \cite{uijlings2013selective} or RPN \cite{ren2015faster}, and then the classifiers will further refine the coordinates of the region proposals proposed in the first stage and simultaneously classify the object inside each bounding box. In contrast, the classifiers of one-stage models directly refine and classify the densely pre-defined anchors. Since the objects to be detected have various shapes and sizes, some object detectors such as SSD \cite{DBLP:journals/corr/LiuAESR15}, FPN \cite{lin2017feature}, RetinaNet \cite{lin2017focal} and RefineDet \cite{zhang2018single} define anchors on various sizes of feature maps so that the anchors in large feature maps can recognize small instances and the anchors in small feature maps can detect large objects. Designing anchors on various sizes of feature maps to detect instances with varying sizes and shapes is so effective that most modern detectors adopt this technique to design network architectures. 

Recently, some anchor-free detectors, like CornerNet \cite{law2018cornernet}, FSAF \cite{zhu2019feature}, ExtremeNet \cite{zhou2019bottom}, CenterNet \cite{duan2019centernet}, and FCOS \cite{tian2019fcos} are designed without any pre-defined anchor boxes. In addition, there is also some state-of-the-art such as \cite{zhang2020bridging} which combines anchor-based detectors and anchor-free detectors to adaptively train the samples.
Two-stage object detectors often outperform one-stage object detectors while one-stage models have faster speed than two-stage models. Since the goal is to detect and track the golf ball in real-time, the detectors chosen have to be both fast and accurate. Most of the one-stage methods have the advantage over two-stage methods in speed with the sacrifice of accuracy. To balance the trade-off, Faster R-CNN \cite{ren2015faster}, YOLOv3 and its light version \cite{Redmon2018YOLOv3AI}, YOLOv4 \cite{bochkovskiy2020yolov4}, SSD \cite{DBLP:journals/corr/LiuAESR15} and RefineDet \cite{zhang2018single} are chosen in our study.

\section{The Proposed Approach}
\begin{figure}[htp]
\centerline{\includegraphics[width=0.5\textwidth]{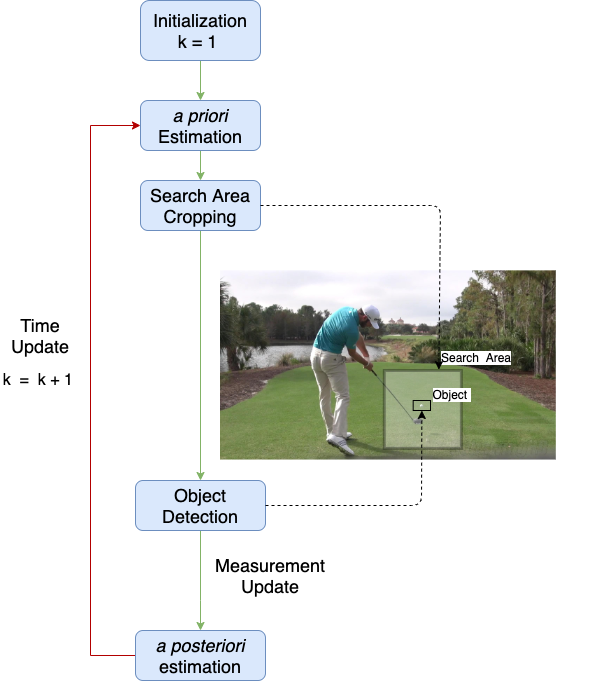}}
\caption{The flow-chart of the proposed tracking strategy. We integrate the object detection and the Kalman filter together for object tracking. In the first frame of the video, we employ the detection model to localize the ball and its coordinates. Each time after the time update, \textit{a priori} state estimate and \textit{a priori} estimate error covariance will be utilized to perform the measurement update. After measurement update, \textit{a posteriori} state estimate and \textit{a posteriori} estimate error covariance will be employed to do the time update to predict \textit{a priori} state estimate and \textit{a priori} estimate error covariance in the next frame.}
\label{process}
\end{figure}

We propose a two-stage scheme for golf ball detection and tracking. As shown in Fig. \ref{process}, we employ a Kalman filter to predict the estimated location of the golf ball in the next consecutive frame. The area centered at the estimated location is cropped and sent to the detector. In general, the image resolution for golf live TV shows is higher than 1080p, which is computationally intensive. To boost the accuracy and reduce the computational loads, we apply the discrete Kalman filter model for the location estimation. The proposed tracking approach depends on a recursive process of \textit{a priori} estimation, object detection, and \textit{a posteriori} estimation. The tracker is initialized in the first frame. In most golf practice driving ranges, the camera is set beside the player so that the full swing is recorded. In this scenario, the initial position of the golf ball is located in the lower center of the frame, which can be easily detected automatically by a detection model. 

For the tracking, after the ball is detected on the cropped image patch, the "Time Update" in the Kalman filter will predict the coordinate of the golf ball in the next frame, which is denoted as \textit{a priori} estimation. The next image will be cropped with respect to the location that the Kalman filter predicts. Then the cropped image patch will be sent to the object detector. After that, based on the detection results, the "Measurement Update" in the Kalman filter will calculate \textit{a posteriori} estimate of the current state which will be sent to the "Time update" in the Kalman filter for \textit{a priori} estimation to predict the coordinate of the ball in the next frame. This loop will continue until all frames are detected.

\subsection{Object Detection}
The detection models we chose in this study are  Faster R-CNN \cite{ren2015faster}, YOLOv3 and its light version \cite{Redmon2018YOLOv3AI}, YOLOv4 \cite{bochkovskiy2020yolov4}, SSD \cite{DBLP:journals/corr/LiuAESR15} and RefineDet \cite{zhang2018single}. Among these models, YOLOv3 \cite{Redmon2018YOLOv3AI}, YOLOv4 \cite{bochkovskiy2020yolov4}, SSD \cite{DBLP:journals/corr/LiuAESR15}, and RefineDet \cite{zhang2018single} are considered as representatives of the one-stage object detection models, while Faster R-CNN \cite{ren2015faster} is a classical two-stage object detector. Object detection is to determine the location of the objects in the images or videos and simultaneously classify those objects in the image or videos.

\paragraph{YOLOv3 \& YOLOv3 Tiny}
YOLOv3 \cite{Redmon2018YOLOv3AI} predicts bounding boxes using pre-defined boxes with dimension clusters. YOLOv3 predicts multiple bounding boxes per grid cell. The prediction is based on the highest IoU between the predicted bounding box and the ground truth bounding box. Non-maximum suppression was applied when multiple bounding boxes appear in the image. The confidence score for the occurrence of the object in each bounding box is predicted using logistic regression. Inspired by the image pyramid, YOLOv3 adds several convolutional layers after the base feature extractor to make predictions at three different scales. Features are extracted from these three scales. YOLOv3 also adds a cross-layer connection between two prediction layers (except the output layer) and earlier finer-grained feature maps. YOLOv3 first up-samples the deep feature maps and then merges it with the previous features by concatenation to better detect small objects.
YOLOv3 provides high accuracy with relatively fast speed. 

YOLOv3 tiny is a simplified version of YOLOv3. It reduced the number of convolutional layers to 7 and utilized $1\times 1$ and $3\times 3$ convolutional layers with fewer parameters. YOLOv3 tiny also utilized a pooling layer to achieve feature map size reduction, while YOLOv3 employed the convolutional layer with a step size of 2.

\paragraph{YOLOv4}
YOLOv4 \cite{bochkovskiy2020yolov4} focuses on a bag of freebies which only changes the training process that might increase the training cost to enhance the performance of the detector without additional cost in inference time and the bag of specials that increases the accuracy of the detector with only a little increase of the inference time by post-processing. The backbone of YOLOv4 is CSPDarknet53 \cite{wang2020cspnet} which is an enhanced version of DarkNet53 used in YOLOv3. To increase the receptive field, YOLOv4 utilizes SPP \cite{he2015spatial} over the backbone network. In addition, instead of employing FPN \cite{lin2017feature} to merge feature maps with different scales in YOLOv3, YOLOv4 exploits PANet \cite{liu2018path} to combine different network levels. The detector head of YOLOv4 is still the same as YOLOv3. There are many other bag of freebies and bag of specials utilized in YOLOv4 to improve the performance of the detector. Some of them would increase little inference cost but improve the performance significantly \cite{bochkovskiy2020yolov4}.

\paragraph{Faster R-CNN}
Faster R-CNN\cite{ren2015faster} is built on top of two networks, region proposal network (RPN) and detection network. RPN is responsible for generating candidate proposals from a series of densely sampled anchors with different sizes and ratios in each location on the feature maps. By default, there are 3 scales and 3 aspect ratios totally 9 anchors at each feature map location. The output of the RPN is a bunch of rectangular proposals that may contain some objects in the image and associate with objectness scores. The proposals will later be processed by the classifiers to verify the occurrence of the objects and the regressors to further refine the bounding boxes of the region proposals. To be more accurate, The function of RPN is to judge if the anchors contain some objects and refine the anchor boxes if they contain some objects. Then the region proposals with a high possibility to be foreground will be sent to the detection network to do the classification and bounding box regression. 

For detection network, Fast R-CNN \cite{DBLP:journals/corr/Girshick15} is adopted. ROI pooling is leveraged to the region proposals that may contain some objects and then two branches are adopted, one branch is for refining the bounding boxes of the region proposals, and the other is to classify the objects that may be contained in the region proposals.
To make this network fast and accurate, RPN shares the full-image convolutional features of all region proposals and by ROI pooling, the features of region proposals can be fixed and easily fed into the detection network. Due to the introduction of RPN, most background proposals are removed and only those proposals with high possibility containing some objects are fed into the detection network, thus the accuracy of Faster R-CNN is often better than that of one-stage object detectors which directly refines and classifies the dense anchors without selecting them. %Nonetheless, because of RPN, the speed of Faster R-CNN is often slower than that of one-stage models.
In the experiment, VGG-16 \cite{simonyan2014very} is selected as the backbone network for Faster R-CNN.

\paragraph{SSD}
SSD \cite{DBLP:journals/corr/LiuAESR15} is abbreviated for Single Shot Multibox Detector which is a one-stage multiple-scale approach. SSD was utilized VGG-16 \cite{simonyan2014very} as the backbone network. However, SSD \cite{DBLP:journals/corr/LiuAESR15} exploits feature maps with various scales to detect objects with different shapes and sizes. Although SSD achieved great performance on large objects, it shows drawbacks in detecting small objects. Lower level feature maps may contain more details while less semantic information, while deep feature maps may have more semantic information but fewer details. To better detect small objects such as golf balls, we have to extract and highlight the feature at a lower level of feature maps. Nonetheless, lacking enough semantic information in those shallower feature maps may prevent us from achieving good detection results for small objects.

\paragraph{RefineDet}
Similar to SSD \cite{DBLP:journals/corr/LiuAESR15}, RefineDet \cite{zhang2018single} also utilizes pre-defined bounding boxes on multi-scale feature maps and objectness scores associating with those boxes to do the object detection. This network mainly consists of three modules: ARM, ODM, and TCB. ARM provides better initialization for regressors and can be treated as RPN in Faster R-CNN \cite{ren2015faster}. ODM whose function is to detect the objects performs as one-stage multiple-scale detector SSD \cite{DBLP:journals/corr/LiuAESR15}. TCB which is similar to the upsampling process in FPN \cite{lin2017feature} connects ARM and ODM by combining the deep feature maps with the shallow feature maps via deconvolution. TCB serves as a bridge between ARM and ODM and it takes the feature maps associated with anchors and the transferred higher-level feature maps to increase detection accuracy. TCB just deconvolves higher-level feature maps to match the size and dimension of the current feature maps and then simply add the current feature map and the higher-level feature maps element-wise. Thus deep feature maps with more semantic information and shallow feature maps with more details can be combined together to enhance the performance of the detector.

Although it has two steps just like the two-stage object detectors, RefineDet is a one-stage detector since it closely combines the two steps without the process in two-stage detectors, such as RoIPooling. In addition, RefineDet utilizes VGG-16 \cite{simonyan2014very} as the backbone which is the same as Faster R-CNN \cite{ren2015faster} and SSD \cite{DBLP:journals/corr/LiuAESR15}. Employing more advanced networks as the backbone may boost the performance, but for convenience, VGG-16 was harnessed in our experiment.

\subsection{Kalman Filter}
The discrete Kalman filter is employed for location estimation. In Kalman filter, a motion state variable is defined as $x=\{a,b,u,v\}$, where $\{a,b\}$ could be the coordinates of the object, and $\{u,v\}$ could be the velocity of the object along the two directions. From \cite{welch1995introduction}, the state at each frame $k$ is using equation \ref{eq1} and the measurement is using equation \ref{eq2}. $w(k)$ and $v(k)$ are prediction noise and measurement noise, respectively, which are assumed normal distribution with covariance $Q$ and $R$. The transition matrix $A \in n \times n$ calculates \textit{a priori} estimate at current state from previous state. $H$ is denoted as the observation matrix. The equations are from \cite{welch1995introduction}.
\begin{equation}
{x}_{k}=A{x}_{k-1}+B{u}_{k-1}+{w}_{k-1} \label{eq1}
\end{equation}
\begin{equation}
{z}_{k}=H{x}_{k}+{v}_{k} \label{eq2}
\end{equation}

\begin{comment}
where the transition matrix $A$ and the observation matrix $H$ in our Kalman model are shown below.

\begin{equation}
A = 
\left({\begin{array}{cccc} 1 & 0 & 1 & 0\\ 0 & 1 & 0 & 1\\ 0 & 0 & 1 & 0 \\ 0 & 0 & 0 & 1\end{array}}\right) \label{eq3}
\end{equation}

\begin{equation}
H = 
\left({\begin{array}{cccc} 1 & 0 & 1 & 0\\ 0 & 1 & 0 & 1\end{array}}\right) \label{eq4}
\end{equation}
\end{comment}

The update of Kalman filter includes two updates: Time Update and Measurement Update, as shown below.

{\bf Time Update:}
\subsubsection{a priori state estimate}
\begin{equation}
\hat{x}^{-}_{k}=A\hat{x}_{k-1}+Bu_{k-1} \label{eq5}
\end{equation}
\subsubsection{a priori estimate error covariance}
\begin{equation}
{P}^{-}_{k}=A{P}_{k-1}A^T+Q\label{eq6}
\end{equation}

{\bf Measurement Update:}
\subsubsection{Kalman Gain}
\begin{equation}
K_{k} = {P}^{-}_{k}H^T(H{P}^{-}_{k}H^T+R)^{-1} \label{eq7}
\end{equation}
\subsubsection{a posteriori state estimate with measurement $z_{k}$}
\begin{equation}
\hat{x}_{k} =\hat{x}^{-}_{k} + K_{k}(z_{k}-H\hat{x}^{-}_{k})  \label{eq8}
\end{equation}
\subsubsection{a posteriori estimate error covariance}
\begin{equation}
P_{k} = (I-K_{k}H){P}^{-}_{k} \label{eq9}
\end{equation}

$\hat{x}^{-}_{k}$ is \textit{a priori} motion state estimate and $\hat{x}_{k}$ is \textit{a posteriori} motion state estimate given the measurement information. ${P}^{-}_{k}$ represents \textit{a priori} estimate error covariance and \textit{a posteriori} estimate error covariance is represented by ${P}_{k}$. We refer to \cite{welch1995introduction} for more details about those equations.

%We integrate the object detection and the Kalman filter together for object tracking. In the first frame of the sequence, we employ the detection model to localize the ball and its coordinates. Each time after the time update, \textit{a priori} state estimate and \textit{a priori} estimate error covariance will be utilized to perform the measurement update. After measurement update, \textit{a posteriori} state estimate and \textit{a posteriori} estimate error covariance will be employed to do the time update to predict \textit{a priori} state estimate and \textit{a priori} estimate error covariance in the next frame. This loop will continue until the end of the video process.

%\subsection{Tracking}
We integrate the object detectors and the Kalman filter together for golf ball tracking. First, we employ the detection model to localize the ball and its coordinates in the cropped first frame, whose center is the ground truth coordinate in the first frame. By utilizing the ground truth coordinate in the first frame, we can make sure that the cropped image of the first frame contains the golf ball. With the detection results in the first frame, the measurement update can be implemented easily. After that, the time update will predict the next coarse coordinate of the golf ball from the results of the previous measurement update. Then we crop the next frame centered at the predicted coarse coordinates. The cropped image is sent to the detector to do the detection again. The process will be repeated until the last frame has been detected.

%For the first frame, we utilized the ground truth coordinates as the original center coordinates of the golf ball to make sure that the cropped first frame contains the golf ball. Then goes the measurement update and time update. Time update predict the next coarse coordinate of the golf ball from the previous measurement update. Starting from the second frame, we employ the detection model to obtain the center coordinates of the golf ball from the "updated search region". Then goes the measurement update and time update. The process will repeat until the last frame being processed.

\section{Dataset}

Deep learning based models need a large collection of data for training and validation, especially for data that are taken from different viewing points, light conditions, and backgrounds. Most of the data are collected from golf tournaments online and the rest are taken by us. The original data are in video format while we converted those videos into image sequences for labeling, training, and testing. The golf videos are in high resolution and taken using slow motion, to reduce motion blur. The whole dataset consists of 2169 golf images, where 1699 images are utilized as the dataset for detection and the rest 470 images are used for tracking. Details of the dataset are shown in Table~\ref{table:1}.

   \begin{table}[htbp]
 \centering
\caption{Dataset split used for training and test}
\begin{tabular}{ p{2cm}|p{2cm}|p{2cm} }
 \hline
   & No. Images & No.Patches\\
 \hline
 Training Set & 1356 & 11030\\
 Test Set & 343  & 2791 \\
 Tracking  Dataset & 470  & 3615 \\
 \hline
\end{tabular}
\label{table:1}
\end{table}

\subsection{Detection Dataset}

\begin{figure}[htp]
\centering
\includegraphics[width=0.485\textwidth]{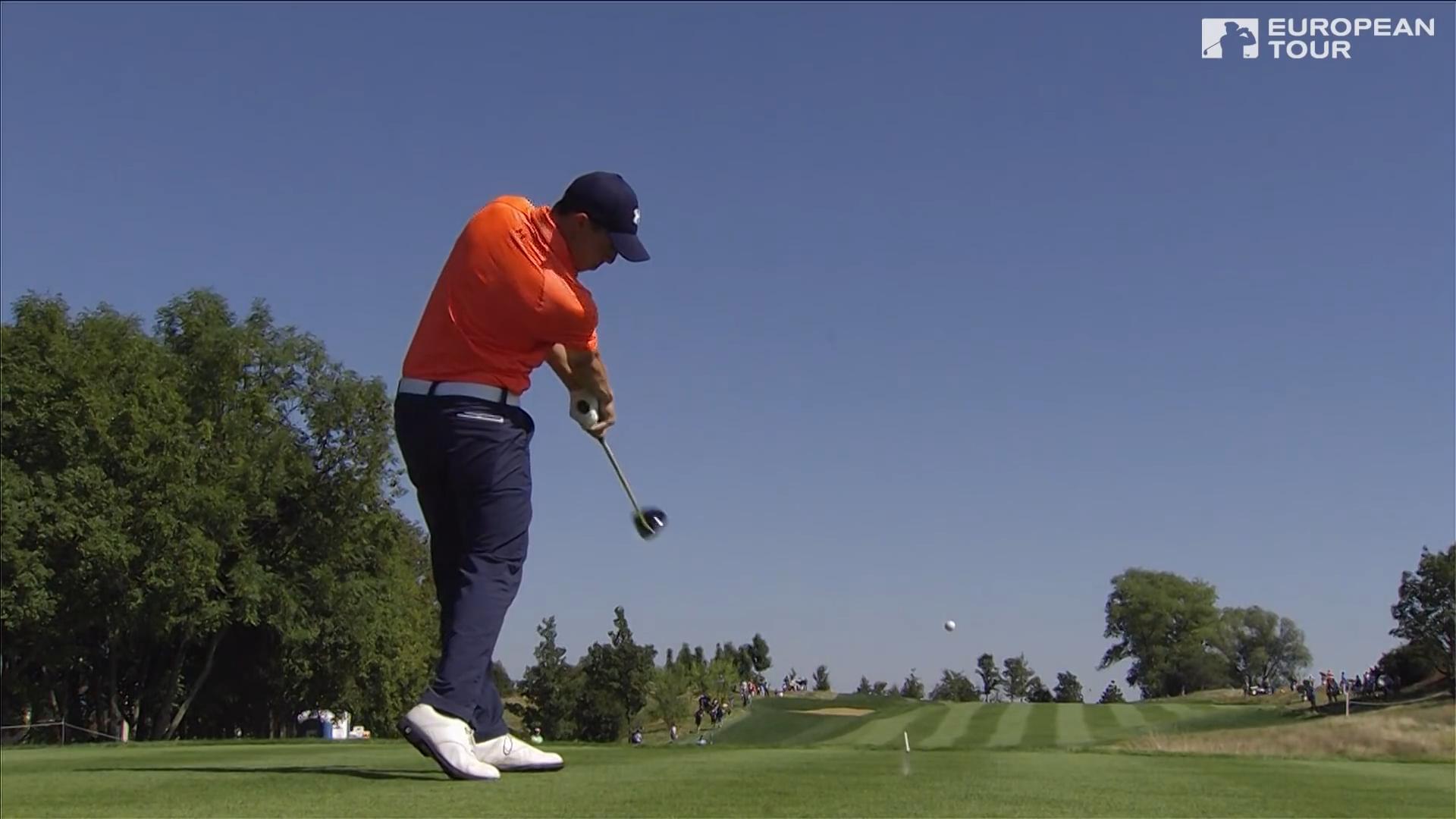}\\
\includegraphics[width=.16\textwidth]{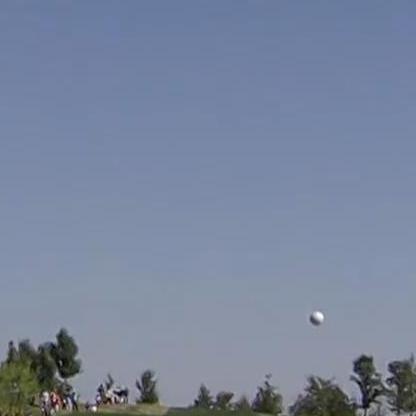}\hfill
\includegraphics[width=.16\textwidth]{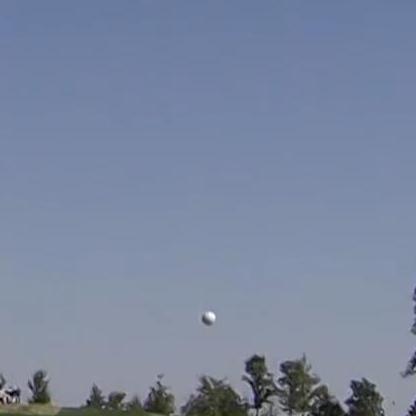}\hfill
\includegraphics[width=.16\textwidth]{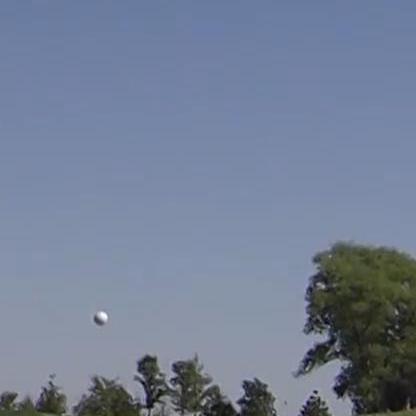}
\includegraphics[width=.16\textwidth]{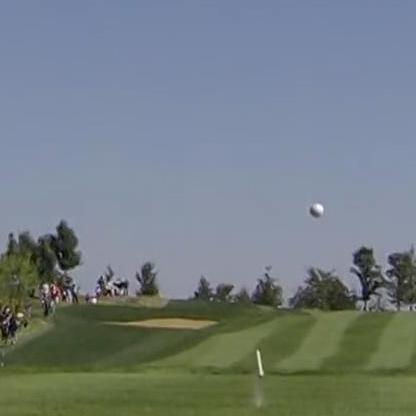}\hfill
\includegraphics[width=.16\textwidth]{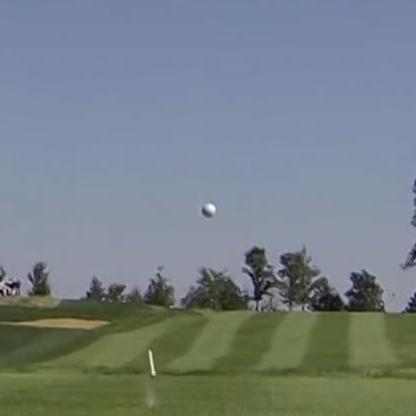}\hfill
\includegraphics[width=.16\textwidth]{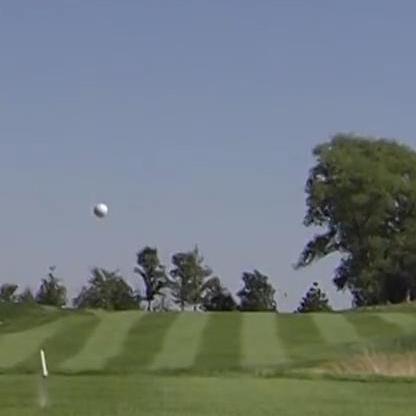}
\includegraphics[width=.16\textwidth]{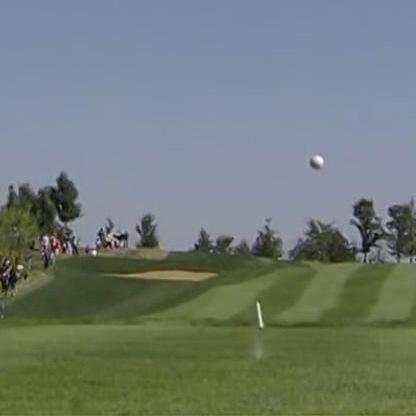}\hfill
\includegraphics[width=.16\textwidth]{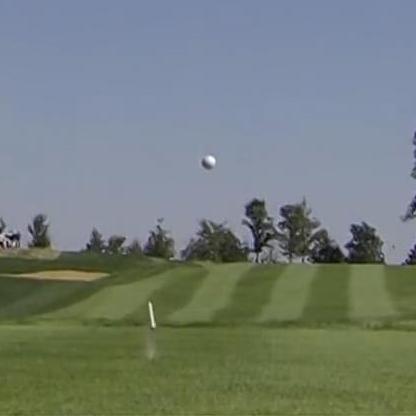}\hfill
\includegraphics[width=.16\textwidth]{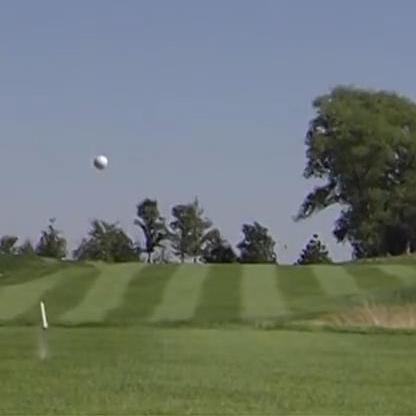}
\caption{An original image and 9 generated training patches. We shift the location to the up, down, left, and right by 100 pixels to generate 9 patches based on the location of the ball. The size of each cropped image patch is $416\times416$.}
\label{fig:figure2}
\end{figure}

To train and validate the detection models, 1699 golf ball images are employed, and a great portion of them are collected from golf tournaments online and the rest are taken by ourselves. We manually label the bounding box of the ball in each image and crop out the $416\times416$ image patch that contains the ball as a training example. Since we only have a limited number of images, we further augment the dataset by shifting the location of the cropped patches. As shown in Fig. \ref{fig:figure2}, based on the location of the ball, we shift the location to the up, down, left, and right by 100 pixels to generate 9 patch samples, with each patch the same size of $416\times416$.  Then, we split the collected images into training and test sets at a ratio of 80:20. The split is based on the original images to ensure the augmented patches in the test set have never been seen in the training set.

%The 9 samples shown in Fig.\ref{fig:figure3} are actually the golf ball being in the middle, shifted to the left, right, up, down, up left, up right, down left and down right by 100 pixels respectively. Once the sample went beyond the boarder, the sample will be eliminated. The non-consecutive image sequences is used for training and testing the detector, while the consecutive image sequences is used for testing and testing the tracker. While training the detectors, image patches generated by the original 1356 images are used for training the detector, the rest 20\% are used for testing. Image size for consecutive image sequences are 1920*1080. Dataset with different background, exposures are collected, cropped and labeled. In our data set, we highlight monocular golf ball case, generally in the golf tournaments, usually one golf ball appears in each frame. We cropped the golf ball from the original image to the size of $416\times416$ rather than compressing the image to have lower resolution. Since the golf ball is tiny in the original frame, compressing the image increases the difficulty in detecting the golf ball. 

\subsection{Tracking Dataset}
In order to test the tracking algorithm and detectors for unseen data, we collected another 17 short swing videos from the golf ball being hit until it flying out of view and 5 golf putting videos. We labeled the ground truth and generated the image patches in the same way as discussed above. We utilize the generated patches to test the generalization ability of the trained detection models on unseen data. The detection results are illustrated in Table~\ref{table:4}. We use the entire video sequence to test the performance of the tracking algorithm. However, some of the short videos have only a few frames, we only choose 8 swing video sequences and 5 putting video sequences with relatively more frames so that we can evaluate the tracking performance, the selected videos and frame numbers are shown in Table~\ref{table:2}. Please note that although the original images in the dataset are of high resolution, the golf balls are very small and only occupy a small number of pixels, most of them are only around $10 \times 10$ pixels, which makes them very hard to detect and track.

\begin{table}[ht]
\centering
\caption{Tracking Dataset}
\begin{tabular}{ |p{3cm}|p{2cm}|p{2cm} |}
 \hline
   & No. Images\\
 \hline
Golf\_Swing\_1 & 19\\
Golf\_Swing\_2 & 33\\
Golf\_Swing\_3 & 16\\
Golf\_Swing\_4 & 28\\
Golf\_Swing\_5 & 13\\
Golf\_Swing\_6 & 35\\
Golf\_Swing\_7 & 22\\
Golf\_Swing\_8 & 97\\
Golf\_Put\_1  & 35\\
Golf\_Put\_2  & 33\\
Golf\_Put\_3  & 25\\
Golf\_Put\_4  & 35\\
Golf\_Put\_5  & 14\\
 \hline
 Total & 405\\
 \hline
\end{tabular}
\label{table:2}
\end{table}

\section{Experimental Evaluations}
In our experiments, YOLOv3 and YOLOv3 tiny are implemented using C++. The rest of the detectors are implemented using Python3 with OpenCV and PyTorch framework. The details are given below:

\begin{itemize}
\item \textbf{Faster R-CNN:} PyTorch, cuDNN, and CUDA
\item \textbf{YOLOv4:} PyTorch, cuDNN, and CUDA
\item \textbf{SSD:} PyTorch, cuDNN, and CUDA
\item \textbf{YOLOv3 \& YOLOv3 tiny:} Darknet, cuDNN, and CUDA
\item \textbf{RefineDet:} PyTorch, cuDNN, and CUDA
\end{itemize}

The experiment is run on a single Nvidia Titan XP GPU, which has a memory of 12 GB. To thoroughly evaluate the performance, we will evaluate our detectors and tracker separately. 

\subsection{Detector Evaluations}
Object detection includes both object localization and object classification. To thoroughly evaluate the performance of an object detector, object localization and classification are utilized as assessments. Commonly used classification evaluation metrics are accuracy, precision, and F1 score. Localization is actually a regression problem. For regression evaluation, MSE (Mean Squared Error) and RMSE (Root Mean Squared Error) are widely adopted. To better evaluate the performance of an object detector, AP score, which assesses both classification accuracy and localization accuracy, was employed in our experiment. The AP metrics in this paper are AP-25 and AP-50. AP-25 represents AP at IoU threshold 0.25 and AP-50 stands for AP at IoU threshold 0.5. AP-50 is the commonly utilized metric for object detection. While in our experiment, we found that with the small size of the golf ball and error in labeling, AP-50 is too strict to evaluate our detectors. Thus AP-25 and AP-50 are both harnessed to evaluate our detectors.

We compared the performance of six classical detection models in the experiment: YOLOv3 \& YOLOv3 tiny \cite{Redmon2018YOLOv3AI}, YOLOv4 \cite{bochkovskiy2020yolov4}, Faster R-CNN \cite{ren2015faster}, SSD \cite{DBLP:journals/corr/LiuAESR15} and RefineDet \cite{zhang2018single}. The performance of these detectors on the test set is shown in Table~\ref{table:3}. It is obvious that Faster R-CNN and YOLOv4 achieve a similar mAP score that is much higher than other detectors. The inference time of Faster R-CNN and YOLOv4 is also very close to each other in this experiment. YOLOv3 tiny achieves the shortest inference time, which is 4 to 12 times shorter than other detectors.

\begin{table}[htbp]
 \centering
\caption{Detection Performance on the Test Set}
\begin{tabular}{ |c|c|c|c|c|}
 \hline
 Method & mAP@.25 & mAP@0.5 & Inference Time & fps \\
 \hline
 Faster R-CNN & 98.3\%  & 95.9\% & 36.00 ms & 27.78\\
 YOLOv4 & 99.3\% & 95.6\% & 35.84 ms & 27.90\\
 YOLOv3 & 95.6\%  & 88.2\% &  17.96 ms & 55.67\\
 YOLOv3 tiny & 92.3\%  & 84.2\% & 2.79 ms & 357.85\\
 SSD & 95.8\%  & 78.3\% & 11.00 ms & 90.91\\
 RefineDet & 90.0\%  & 81.5\% & 20.00 ms & 50\\
 \hline
\end{tabular}
\label{table:3}
\end{table}

In order to test the generalization ability of our trained detection models for new data, we performed the same detection experiments on the image patches generated by all tracking sequences. The results are shown in Table~\ref{table:4}. It is evident that YOLOv4 achieves the best precision in terms of mAP score, while the result of Faster R-CNN is comparable to YOLOv4. However, the performance of all models drops by a large margin compared to the results from the test set. This is mainly due to two reasons: (1) the original training set is relatively small and training samples do not have sufficient representativeness; and (2) the distributions of the video dataset and the original training set are different because they are collected from different sources. Nonetheless, the overall performance is still acceptable and this can be demonstrated from the tracking performance. The inference time and detection speed of the tracking dataset remain the same as the test set as we are using the same size of image patches.

\begin{table}[htbp]
 \centering
\caption{Detector Performance on Tracking Dataset}
\begin{tabular}{ |c|c|c|c|c|}
 \hline
 Method & mAP@.25 & mAP@0.5  \\
 \hline
 Faster R-CNN & 89.0\%  & 84.0\% \\
 YOLOv4 & 92.0\% & 84.4\% \\
 YOLOv3 & 82.3\%  & 71.4\% \\
 YOLOv3 tiny & 83.4\%  & 63.7\% \\
 SSD & 84.8\%  & 65.0\% \\
 RefineDet & 82.0\%  & 64.5\% \\
 \hline
\end{tabular}
\label{table:4}
\end{table}

\begin{comment}
\begin{table}[htbp]
 \centering
\caption{Detector Performance on Tracking Dataset}
\begin{tabular}{ |c|c|c|c|c|}
 \hline
 Method & mAP@.25 & mAP@0.5  \\
 \hline
 Faster R-CNN & 89.3\%  & 84.4\% \\
 YOLOv4 & 92.3\% & 84.2\% \\
 YOLOv3 & \%  & \% \\
 YOLOv3 tiny & \%  & \% \\
 SSD & 85.0\%  & 63.7\% \\
 RefineDet & 81.5\%  & 63.3\% \\
 \hline
\end{tabular}
\label{table:4}
\end{table}
\end{comment}

\begin{comment}
\begin{table}[htbp]
 \centering
\caption{Detector Performance on Tracking Dataset}
\begin{tabular}{ |c|c|c|c|c|}
 \hline
 Method & mAP@.25 & mAP@0.5 & Inference Time & fps \\
 \hline
 Faster R-CNN & 89.0\%  & 84.0\% &  36.00 ms & 27.78\\
 YOLOv3 & 82.3\%  & 71.4\% & 17.96 ms & 55.67\\
 YOLOv3 tiny & 83.4\%  & 63.7\% & 2.79 ms & 357.85\\
 SSD & 84.8\%  & 65.0\% & 12.55 ms & 79.68\\
 RefineDet & 82.0\%  & 64.5\% & 20 ms & 50\\
 \hline
\end{tabular}
\label{table:4}
\end{table}
\end{comment}

\subsection{Tracking Evaluation}
We adopted the evaluation metrics from \cite{WuLimYang13} to evaluate our tracking results. We use the selected 13 sequences in Table~\ref{table:2} to evaluate our algorithm. The results are evaluated in terms of CLE (center location error), precision, and SR (success rate). CLE is calculated by the Euclidean distance between the center location of the tracked object and the ground truth. Precision describes the percentage of the images whose predicted location is within a certain threshold of the ground truth location. 
%Table~\ref{table:5} shows the precision results of all 5 models with CLE being 1, 2, and 5. It is obvious that Faster R-CNN still has the best performance under CLE metrics. %Fig.\ref{Precision Plot} shows the overall precision of our tracker with threshold varied from 0 to 1. 

%\begin{figure}[htbp]
%\centerline{\includegraphics[width=0.5\textwidth]{precision.png}}
%\caption{Precision Plot}
%\label{Precision Plot}
%\end{figure}
\begin{table*}[htbp]
\centering
\caption{Tracker Precision Rate Evaluation}
\resizebox{\textwidth}{!}{
\begin{tabular}{ |c|c|c|c|c|c|c|c|c|c|c|c|c|c|c|c|}
 \hline
 \multirow{2}{*}{Sequence} & \multicolumn{3}{c|}{Faster R-CNN} & 
     \multicolumn{3}{c|}{YOLOv3} & \multicolumn{3}{c|}{YOLOv3 tiny} & 
     \multicolumn{3}{c|}{SSD} & \multicolumn{3}{c|}{RefineDet} \\
 \cline{2-16}
 & CLE\_1 & CLE\_2 & CLE\_5 & CLE\_1 & CLE\_2 & CLE\_5 & CLE\_1 & CLE\_2 & CLE\_5 & CLE\_1 & CLE\_2 & CLE\_5 & CLE\_1 & CLE\_2 & CLE\_5 \\ 
 \hline
Golf\_Swing\_1 & 31.6\% & 63.2\% & 89.5\% & 15.8\% & 36.8\% & 68.4\% & 15.8\% & 52.6\% & 68.4\% & 26.3\% & 57.9\% & 84.2\% & 5.3\% & 10.5\% & 21.1\% \\

Golf\_Swing\_2 & 36.4\% & 75.8\% & 93.9\% & 30.3\% & 45.5\% & 90.9\% & 24.2\% & 48.5\% & 81.8\% & 15.2\% & 48.5\% & 90.9\% & 36.4\% & 78.8\% & 90.9\% \\

Golf\_Swing\_3 & 31.3\% & 75.0\% & 81.3\% & 0\% & 18.8\% & 25.0\% & 31.3\% & 56.3\% & 56.3\% & 6.3\% & 6.3\% & 43.8\% & 12.5\% & 18.8\% & 43.8\% \\

Golf\_Swing\_4 & 17.9\% & 32.1\% & 50.0\% & 7.1\% & 25.0\% & 57.1\% & 3.6\% & 21.4\% & 57.1\% & 0\% & 0\% & 0\% & 7.1\% & 21.4\% & 78.6\% \\

Golf\_Swing\_5 & 15.4\% & 30.8\% & 61.5\% & 23.1\% & 46.2\% & 76.9\% & 0\% & 38.5\% & 92.3\% & 0\% & 30.8\% & 92.3\% & 7.7\% & 23.1\% & 53.9\% \\

Golf\_Swing\_6 & 37.1\% & 65.7\% & 97.1\% & 0\% & 20.6\% & 82.4\% & 14.7\% & 29.4\% & 64.7\% & 5.7\% & 31.4\% & 88.6\% & 11.4\% & 34.3\% & 94.3\% \\

Golf\_Swing\_7 & 36.4\% & 95.5\% & 100.0\% & 4.76\% & 33.3\% & 71.4\% & 23.8\% & 57.1\% & 95.2\% & 0\% & 0\% & 0\% & 22.7\% & 50.0\% & 86.4\% \\

Golf\_Swing\_8 & 56.7\% & 81.4\% & 100.0\% & 17.5\% & 50.5\% & 92.8\% & 10.3\% & 39.2\% & 85.6\% & 9.3\% & 33.0\% & 97.9\% & 38.1\% & 68.0\% & 87.6\%\\

Golf\_Put\_1 & 34.3\% & 82.9\% & 91.4\% & 2.9\% & 37.1 \% & 94.3\% & 5.71\% & 31.4\% & 88.6\% & 0\% & 8.6\% & 94.3\% & 2.9\% & 11.4\% & 94.3\% \\

Golf\_Put\_2 & 36.4\% & 78.8\% & 100.0\% & 27.3\% & 66.7\% & 100.0\% & 12.1\% & 54.6\% & 97.0\% & 0\% & 9.1\% & 97.0\% & 6.1\% & 21.2\% & 97.0\%\\

Golf\_Put\_3 & 36.0\% & 80.0 \% & 100.0\% & 12.4\% &  40.0\% & 100.0\% & 8.0\% & 32.0\% & 96.0\% & 0\% & 0\% & 96.0\% & 0\% & 32.0\% & 88.0\% \\

Golf\_Put\_4 & 22.9\% & 71.4 \% & 100.0\% & 17.1\% & 42.9\% & 97.1\% & 2.9\% & 37.1\% & 91.4\% & 2.9\% & 11.4\% & 97.1\% & 5.7\% & 14.3\% & 85.7\% \\

Golf\_Put\_5 & 57.1\% & 85.7\% & 100.0\% & 7.14\% & 42.9\% & 100.0\% & 14.3\% & 35.7\% & 92.9\% & 0\% & 14.3\% & 78.6\% & 7.1\% & 21.4\% & 78.6\% \\

 \hline
Average & 34.6\% & 70.1\% & 89.6\% & 12.7\% & 38.9\% & 81.3\% & 12.8\% & 41.1\% & 82.1\% & 5.0\% & 19.3\% & 73.9\% & 12.5\% & 31.2\% & 76.9\%\\
 \hline
 \end{tabular}
 }
 \label{table:5}
 \end{table*}

Another evaluation metric is the bounding box overlap. Tracking result is considered successful if $|\frac{{x}_{t}\cap {x}{g}}{{x}_{t}\cup {x}_{g}}|>\theta$. $\theta \in \left[ 0,1 \right]$. ${x}_{t}$ is denoted as the tracking bounding box, while ${x}_{g}$ is denoted as ground truth bounding box. The number of successful frames are counted in order to measure the performance of a sequence. Rather than assigning a certain threshold, we swept the threshold from 0 to 1 and used AUC (area under the curve) to evaluate the tracker. 
%Table~\ref{table:6} demonstrates the success rate of our trackers on all sequences. The overall success rate and precision rate of all 5 models is depicted in Fig. \ref{success Plot} and Fig. \ref{precision Plot}. According to Fig. \ref{success Plot} and Fig. \ref{precision Plot}, Faster R-CNN still has the best performance for success rate and precision rate by any overlap threshold and center location error threshold. The best model among the other 4 models varies when the threshold changes.

\begin{table}[htbp]
\centering
\caption{Tracker Success Rate Evaluation}
\resizebox{\columnwidth}{!}{
\begin{tabular}{ |c|c|c|c|c|c|}
 \hline
Sequence & Faster R-CNN & YOLOv3 & YOLOv3 tiny & SSD & RefineDet\\
 \hline
Golf\_Swing\_1 & 54.0\% & 39.4\% & 36.8\% & 43.3\% & 15.0\% \\
Golf\_Swing\_2 & 61.2\% & 48.2\% & 45.0\% & 54.0\% & 60.9\% \\
Golf\_Swing\_3 & 55.6\% & 15.5\% & 36.7\% & 23.8\% & 27.2\% \\
Golf\_Swing\_4 & 29.2\% & 27.4\% & 26.1\% & 0.0\% & 39.5\%  \\
Golf\_Swing\_5 & 41.8\% & 47.3\% & 51.5\% & 54.6\% & 36.8\% \\
Golf\_Swing\_6 & 73.4\% & 52.0\% & 47.2\% & 55.8\% & 63.1\% \\
Golf\_Swing\_7 & 67.1\% & 43.6\% & 55.4\% & 0.0\% & 58.9\% \\
Golf\_Swing\_8 & 77.1\% & 60.9\% & 52.0\% & 61.0\% & 67.7\% \\
Golf\_Put\_1 & 68.3\% & 62.2\% & 61.9\% & 56.2\% & 61.1\% \\
Golf\_Put\_2 & 74.7\% & 72.4\% & 69.1\% & 54.1\% & 63.7\% \\
Golf\_Put\_3 & 83.6\% & 73.9\% & 71.7\% & 63.9\% & 68.0\% \\
Golf\_Put\_4 & 77.6\% & 71.4\% & 66.1\% & 62.2\% & 61.1\% \\
Golf\_Put\_5 & 70.4\% & 67.9\% & 67.6\% & 58.4\% & 61.4\% \\
\hline
Average & 64.3\% & 52.5\% & 52.9\% & 41.8\% & 52.6\% \\
 \hline
\end{tabular}
 }
 \label{table:6}
 \end{table}

\begin{figure}[htbp]
\centerline{\includegraphics[width=0.45\textwidth]{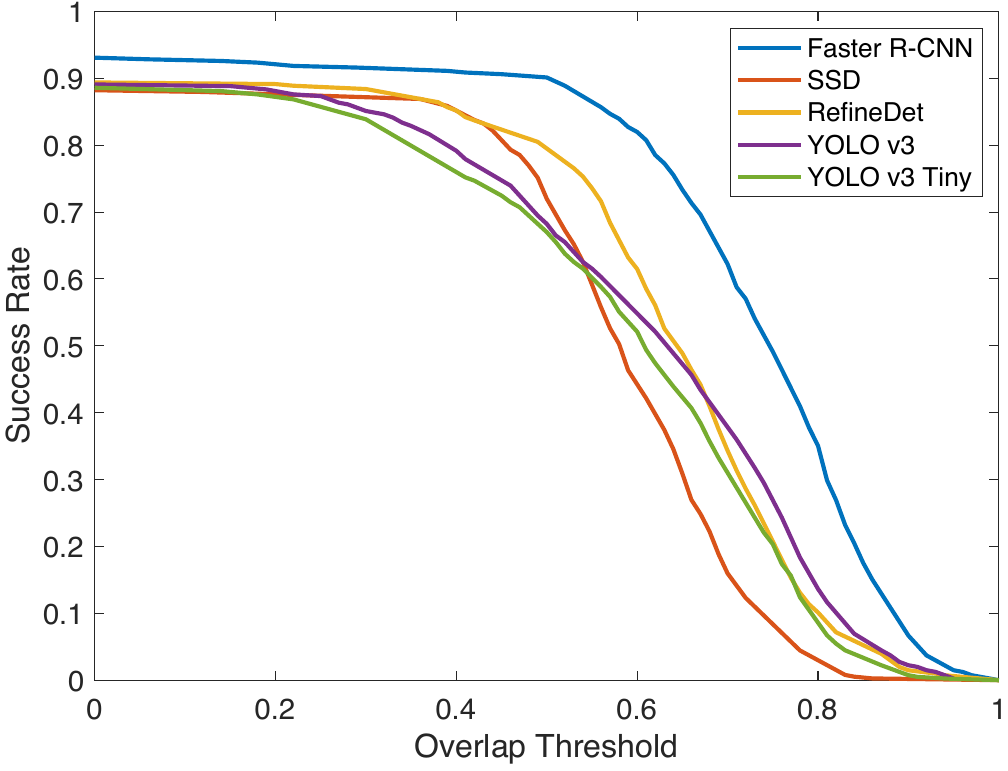}}
 \caption{Success Rate. When the overlap of the ground truth bounding box and the detected bounding box is larger than some threshold in one frame, the tracking result in this frame is denoted as a success. When the overlap threshold increases, which means the accuracy requirement is higher, the success rate drops.}
\label{success Plot}
\end{figure}

\begin{figure}[htbp]
\centerline{\includegraphics[width=0.45\textwidth]{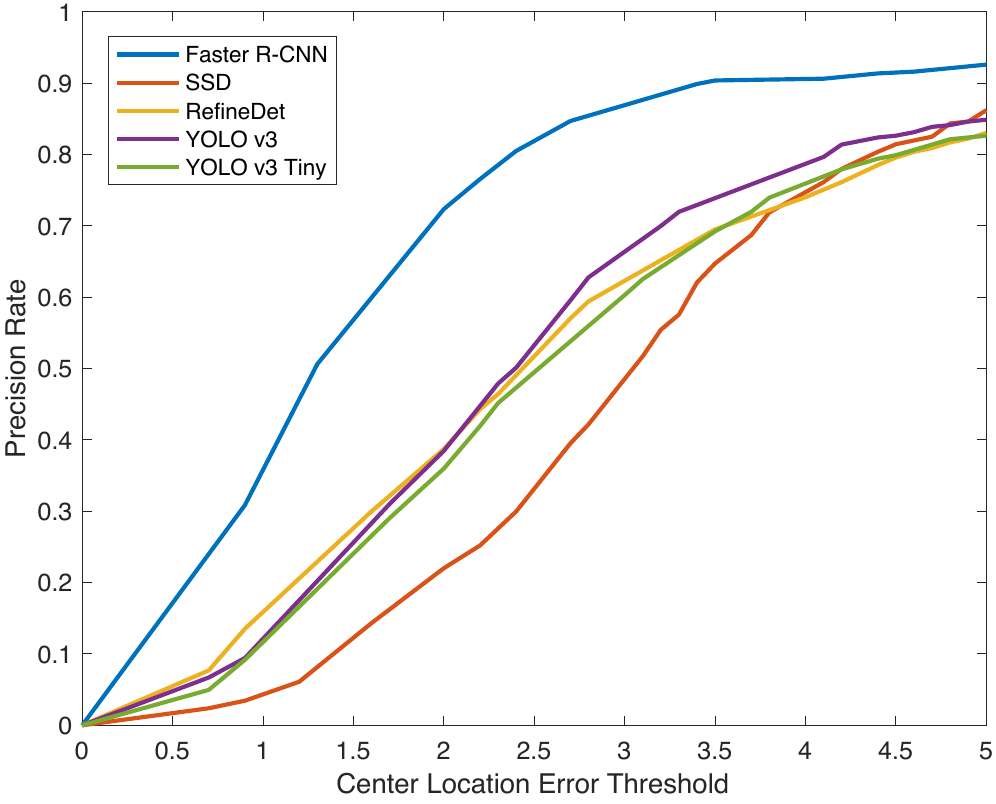}}
\caption{Precision Rate. Low CLE represents the overlap between the detected bounding box and the ground truth bounding box is high, which indicates high accuracy of the detection result. The precision rate increases as the CLE threshold increases.}
\label{precision Plot}
\end{figure}

\subsection{Tracking Results}
%To better visualize the result, the tracked golf ball is superimposed onto the reference frame to show the trajectory of the tracked ball.  Fig.\ref{Tracking Result 1} shows the tracking  results of Golf\_6 and Golf\_7 video sequences, respectively. We can see the tracking result is visually very accurate, giving the fact that most classical tracking algorithms fail to track the ball due to its small sizes and fast movement. %demonstrated great performance of our tracker which is in correspondence with the result obtained using success rate metric. 

%Although our tracking system can achieve demonstrated great frame rate and accuracy, it does have 3 drawbacks to be improved. First, it needs to be initialized in the first frame manually. Second, once the object is moving too fast in image coordinates, generally means, moving faster than 208 pixels per frame. Then the golf ball will go beyond the search region and led to a failure result. Third, the models chosen are light-weighted, while to realize real time, GPU is still a must.

Since the detection results and the inference time of Faster R-CNN \cite{ren2015faster} and YOLOv4 \cite{bochkovskiy2020yolov4} are similar, we only compare the tracking results based on YOLOv3 and YOLOv3 tiny \cite{Redmon2018YOLOv3AI}, Faster R-CNN \cite{ren2015faster}, SSD \cite{DBLP:journals/corr/LiuAESR15}, and RefineDet \cite{zhang2018single} in this experiment. Table~\ref{table:5} shows the precision results of all 5 models with CLE being 1, 2, and 5, respectively. It is obvious that Faster R-CNN has the best performance under the CLE metrics. Table~\ref{table:6} demonstrates the success rate of the trackers on all sequences. The overall success rate and precision rate of 5 models are depicted in Fig. \ref{success Plot} and Fig. \ref{precision Plot}, respectively. Fig. \ref{success Plot} and Fig. \ref{precision Plot} demonstrate the performance of the trackers using the above evaluation metrics. According to Fig. \ref{success Plot} and Fig. \ref{precision Plot}, Faster R-CNN has the best performance for success rate and precision rate at all overlap threshold and center location error threshold. %The best model among the other 4 models varies when the threshold changes. 
We can also see that the two-stage Faster R-CNN achieves the dominant performance in terms of the mAP score, while SSD does not seem to perform well. The performance of YOLOv3, YOLOv3 tiny, and RefineDet is similar. However, YOLOv3 and YOLOv3 tiny, SSD and RefineDet achieve competing accuracy with a much shorter inference time. The best-buy detector is YOLO v3 tiny, which achieved the highest speed of near 358 frames per second (fps) with reasonable accuracy. 
%Fig. \ref{tracking property} visualizes the inference rate for object detection versus AP@0.5 for all 5 models on the detection dataset. It is evident that Faster R-CNN has the best AP performance but the lowest frame rate, while YOLOv3 tiny achieves the highest inference rate with acceptable AP@0.5 value.

Table~\ref{table:7} demonstrates the real frame rate of Faster R-CNN, SSD, and RefineDet in fps (frames per second) of different tracking sequences. Since YOLOv3 and YOLOv3 tiny are written in C++ using different computing platform, we did not test their speed in this experiment. However, based on the detection speed, we infer the fps of YOLOv3 would be in the range of 40-55, and the fps of YOLOv3 tiny could be in the range of 100-180, making it an excellent model for real-world applications.

\begin{comment}

\begin{figure}[tbp]
\centerline{\includegraphics[width=0.5\textwidth]{success_detail.png}}
\caption{Success Rate Plot}
\label{Success Rate Plot}
\end{figure}

\begin{figure}[tbp]
\centerline{\includegraphics[width=0.5\textwidth]{precision_detail.png}}
\caption{Precision Rate Plot}
\label{Precision Rate Plot}
\end{figure}

\begin{figure}[tbp]
\centerline{\includegraphics[width=0.5\textwidth]{Tracking_property.png}}
\caption{Detection speed versus AP@0.5 on the detection dataset. It is noticeable that Faster R-CNN has the best average precision and YOLOv3 tiny yields the highest frame rate.}
\label{tracking property}
\end{figure}
\end{comment}

%\begin{figure}[tbp]
%\centerline{\includegraphics[width=0.5\textwidth]{Detection_Property.png}}
%\caption{Detection speed versus AP@0.5 on the tracking dataset. Faster R-CNN has the highest %average precision among other detectors while YOLO v3 tiny has the shortest inference time.}
%\label{detection property}
%\end{figure}

\begin{table}[htbp]
\centering
\caption{Tracker Frame Rate Evaluation (FPS)}
\centering
\begin{tabular}{ |c|c|c|c|}
 \hline
Sequence & Faster R-CNN & SSD & RefineDet\\
 \hline
Golf\_Swing\_1  & 24.27 & 42.30 & 38.78 \\
Golf\_Swing\_2  & 24.93 & 49.65 & 43.84 \\
Golf\_Swing\_3  & 24.94 & 46.80 & 34.22 \\
Golf\_Swing\_4  & 20.63 & 46.16 & 30.83 \\
Golf\_Swing\_5  & 24.10 & 43.22 & 38.79 \\
Golf\_Swing\_6  & 24.83 & 38.19 & 43.91 \\
Golf\_Swing\_7  & 24.33 & 34.98 & 42.00 \\
Golf\_Swing\_8  & 25.64 & 49.76 & 43.49 \\
Golf\_Put\_1  & 25.58 & 53.40 & 44.54 \\
Golf\_Put\_2  & 26.18 & 56.61 & 46.42 \\
Golf\_Put\_3  & 26.04 & 48.32 & 44.06 \\
Golf\_Put\_4  & 26.39 & 56.03 & 45.34 \\
Golf\_Put\_5  & 25.84 & 42.74 & 46.58 \\
\hline
Average & 24.90 & 46.78 & 41.75 \\
 \hline
\end{tabular}
 \label{table:7}
 \end{table}

\begin{figure}[htbp]
\centerline{\includegraphics[width=0.5\textwidth]{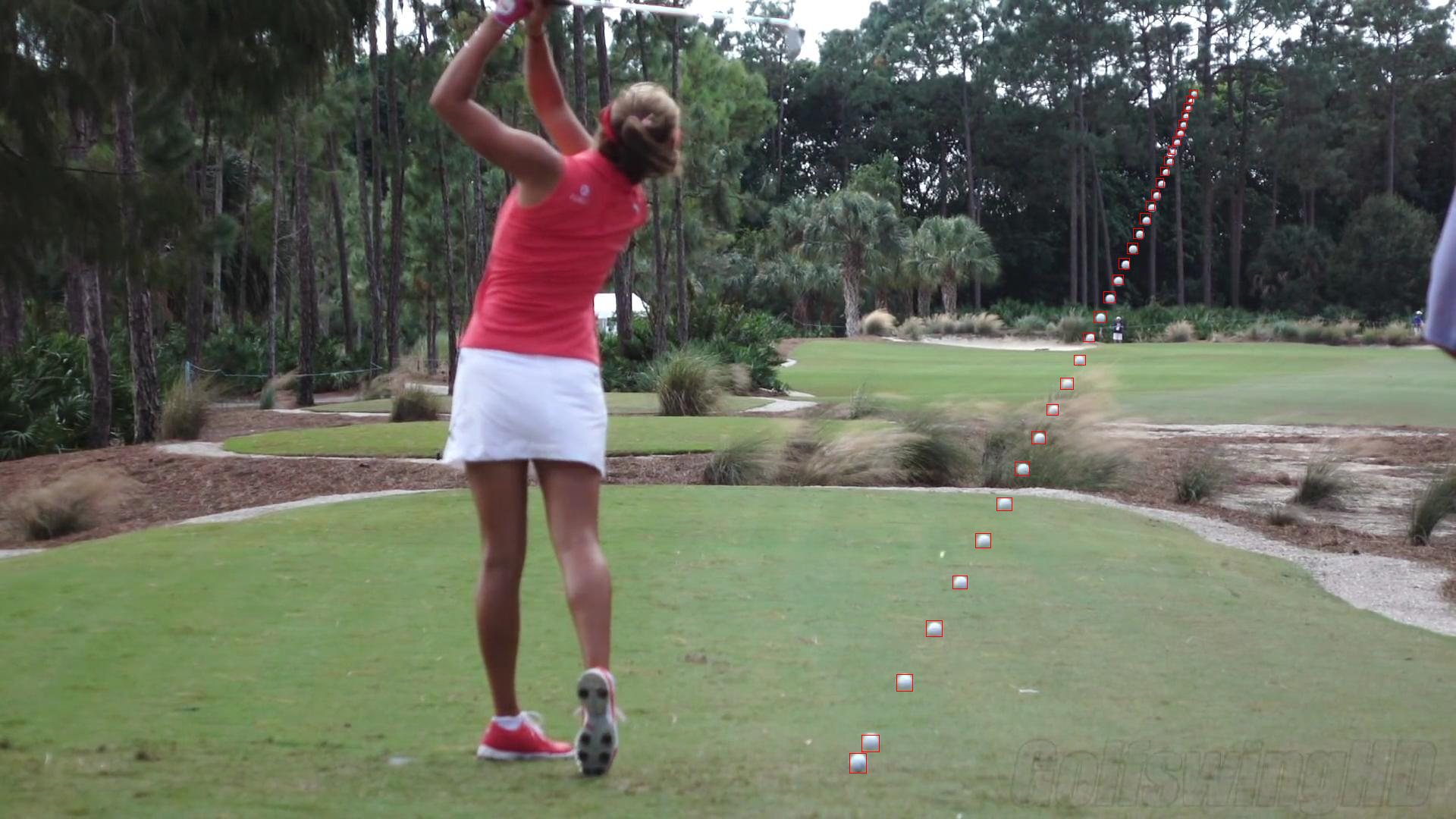}}
\centerline{\includegraphics[width=0.5\textwidth]{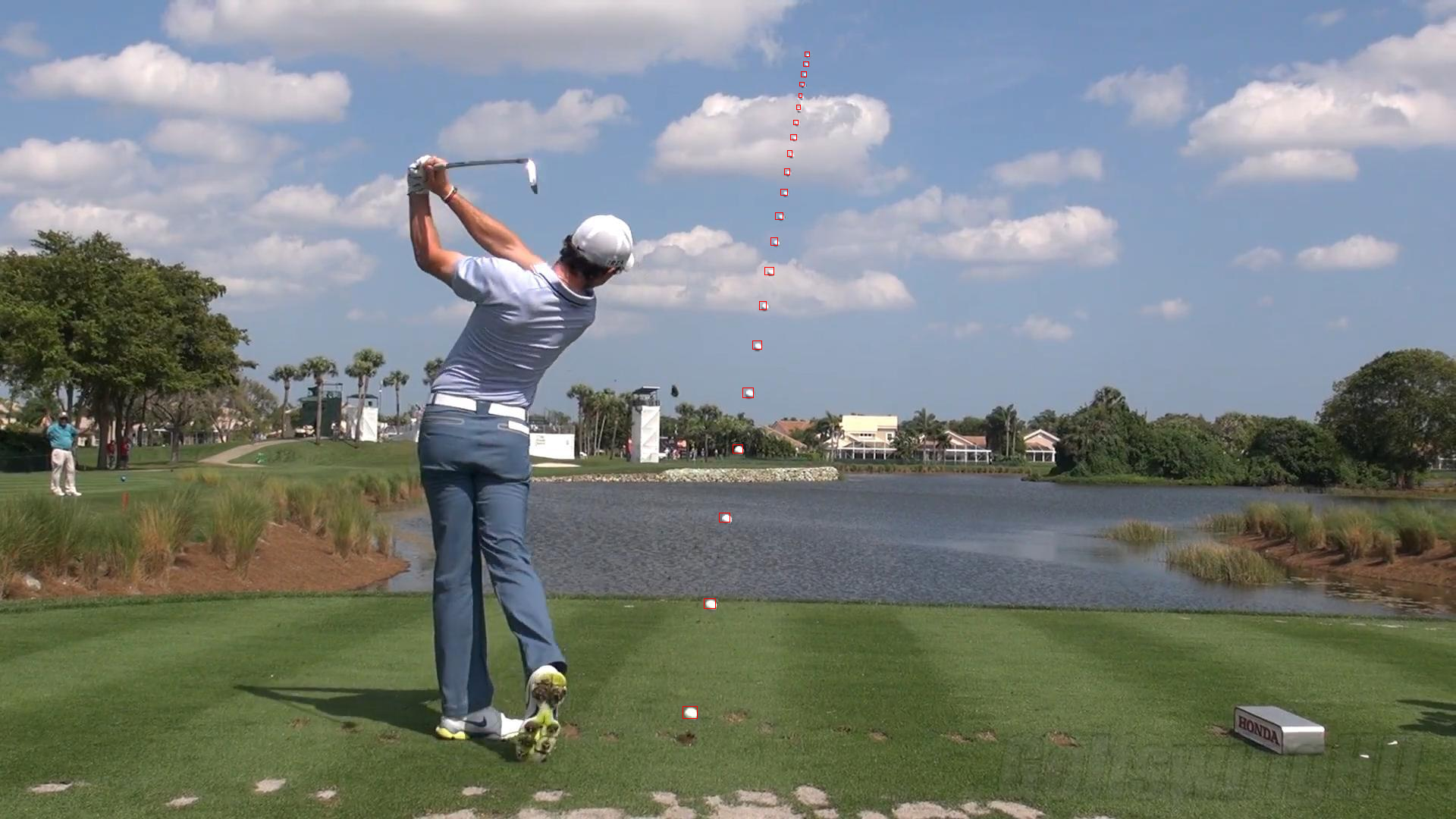}}
\caption{Two sample tracking results based on Faster R-CNN for swing test sequences Golf\_6 (upper) and Golf\_7 (lower). The white golf ball indicates the ground truth, and the red bounding boxes represent the tracking results.}
\label{Tracking Result 1}
\end{figure}

\begin{figure}[htbp]
\centerline{\includegraphics[width=0.5\textwidth]{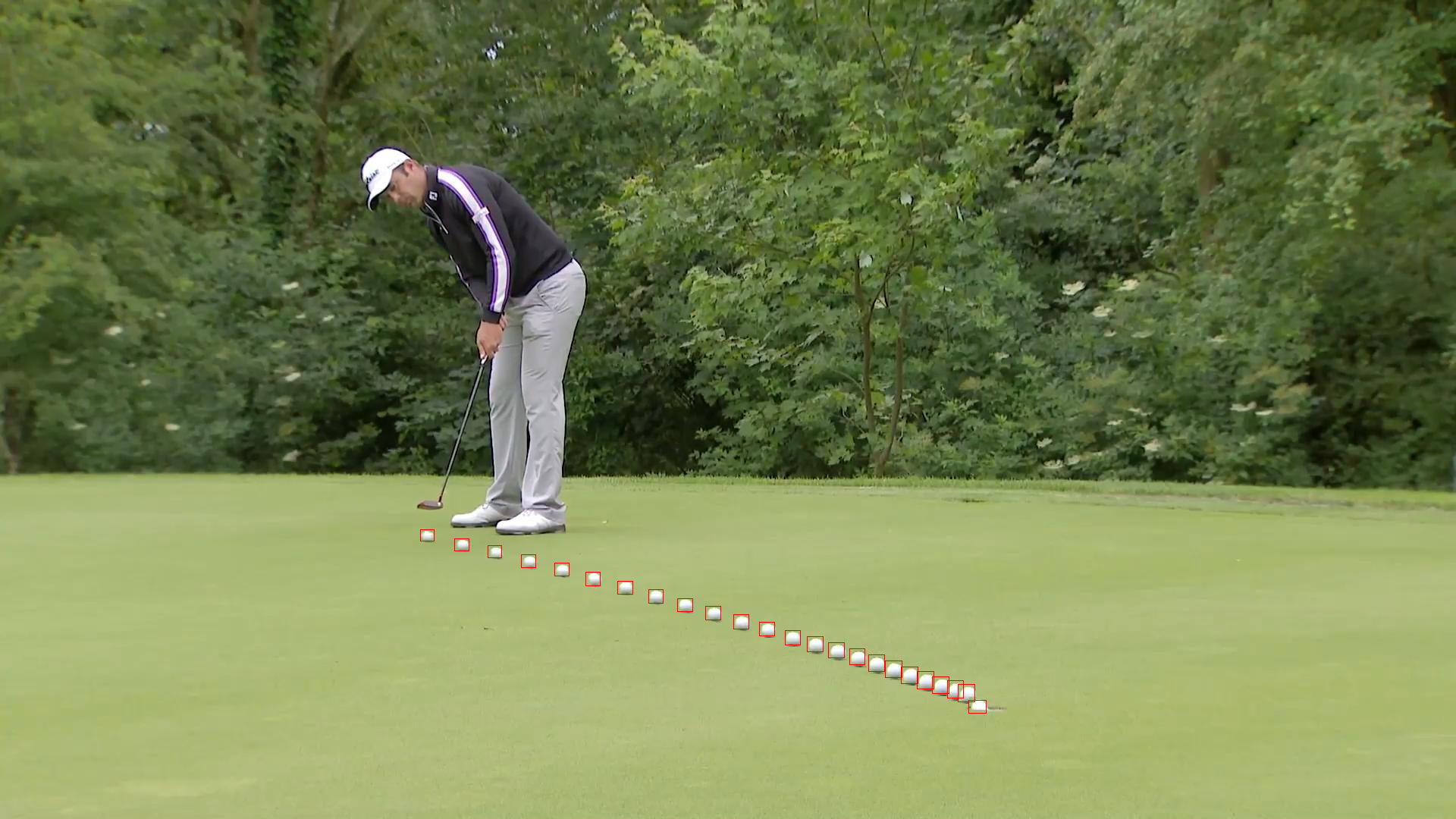}}
\centerline{\includegraphics[width=0.5\textwidth]{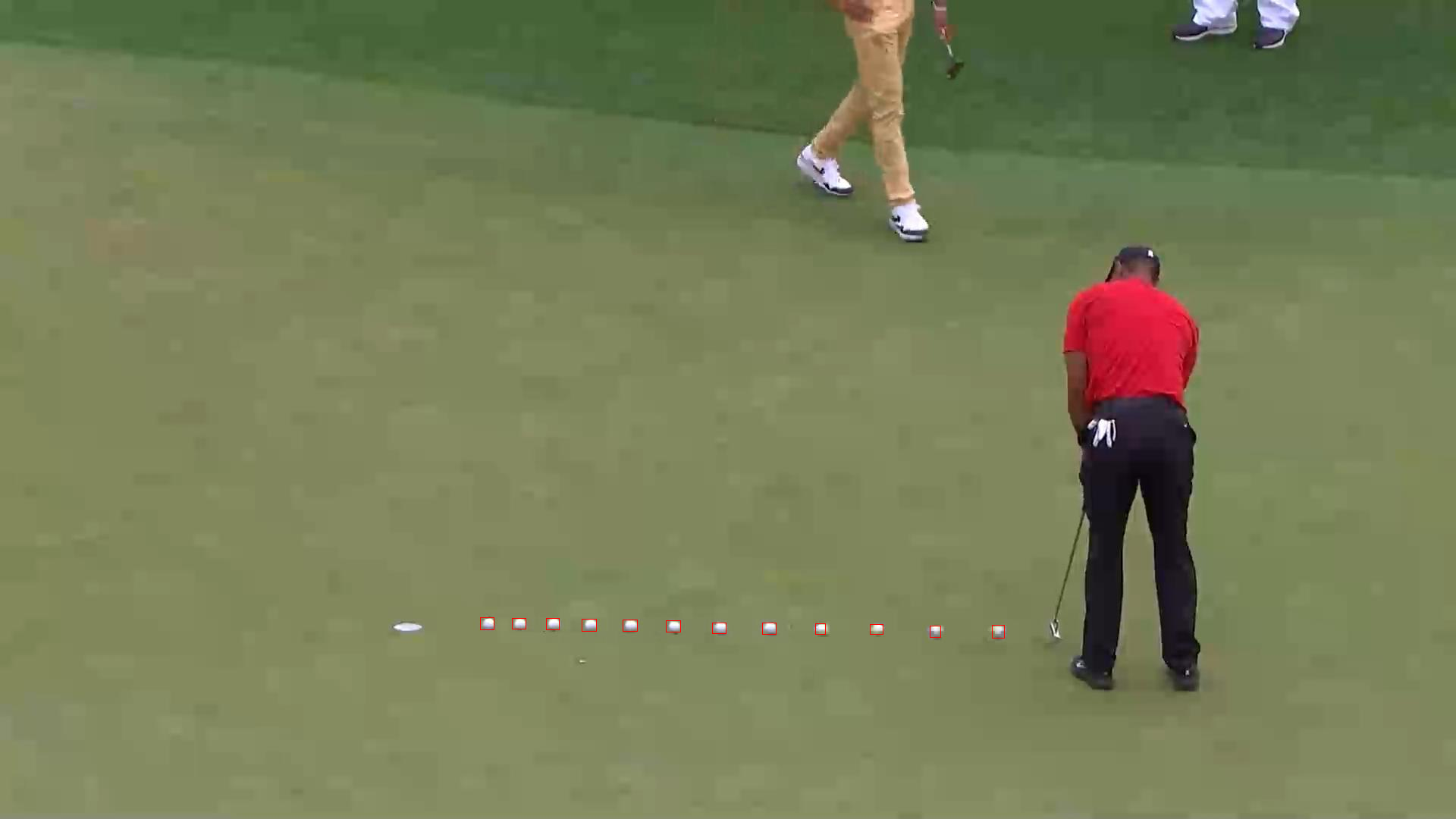}}
\caption{Two sample tracking results based on Faster R-CNN for putting test sequences Put\_3 (upper) and Put\_5 (lower). The white golf ball indicates the ground truth, and the red bounding boxes represent the tracking results. }
\label{Tracking Result 2}
\end{figure}

Fig. \ref{Tracking Result 1} shows the tracking results with Faster R-CNN of Golf\_6 and Golf\_7 video sequences, respectively. Fig. \ref{Tracking Result 2} illustrates the tracking results using Faster R-CNN of two putting sequences. We can see that the tracking results are visually very accurate. While most classical tracking algorithms based on the entire frames fail to track the ball due to the small size and fast movement of the ball.

\section{Experiment and Error Analysis}
In the previous report, most object detectors do not perform well on small object detection. One reason is that most detectors are trained on the datasets, like PASCAL VOC \cite{everingham2015pascal}\cite{everingham2010pascal} or COCO \cite{lin2014microsoft}, where the size of the objects varies greatly, including plenty of large objects and middle-sized objects. The detectors are trained to detect objects of different sizes, rather than solely focusing on small objects. Therefore, the performance of small object detection is not appealing.

In this paper, we only focus on detecting one small object-the golf ball. Thus, we can customize the detectors for this specific task. For instance, the anchor scales in the original Faster R-CNN paper are 128, 256, and 512, while we change them to 8, 16, and 32 in our experiment to better accommodate the small balls in the dataset. In addition, we perform the detection on the cropped patches instead of on the original high-resolution image, which makes the ball relatively ``larger" and easier to detect than taking the entire image as input. To ensure the cropped patch contains the ball, we employ the Kalman filter to predict the location of the ball. As a result, we have achieved relatively good results in both detection and tracking.

%\section{Error Analysis}
The golf ball itself is an extremely small object. As it flies away from the camera, it becomes smaller and smaller, which results in significant detection errors. In addition, since the ball only occupies a small area, it is hard to label it accurately, and one-pixel annotation error may have a huge influence on the precision evaluation of the detection results. 

\begin{figure}[htbp]
\centerline{\includegraphics[width=0.5\textwidth]{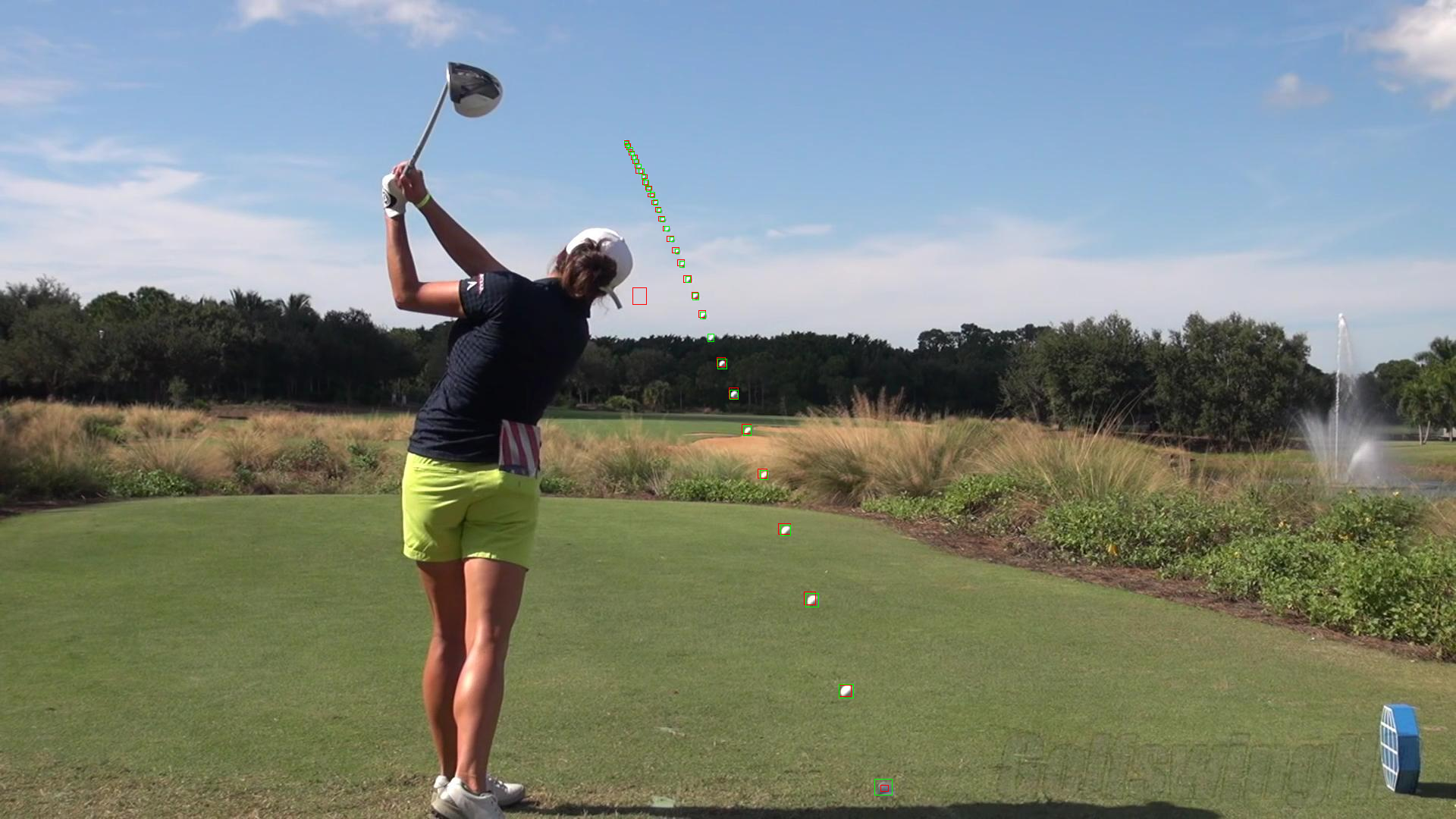}}
\caption{The ground truth and the superimposed tracking results. The green bounding boxes are the ground truth and the red bounding boxes are the tracking results. Please note that there is one falsely tracked ball position.}
\label{Golf Ball Trace}
\end{figure}

As in the example shown in Fig. \ref{Golf Ball Trace}, where the labeled ground truth is represented by the green bounding box, while the tracked results are in red. The size of the golf ball in the first frame is around $27\times27$. It was labeled as $26\times26$ with only 1-pixel error. Suppose the prediction is perfect, while the 1-pixel labeling error could lower the IOU by around 7.5\%. When the ball is away from the camera, its size will decrease to $6\times6$ pixels. In this case, the 1-pixel error could result in 33\% IOU drop. Based on this analysis, we can see that labeling errors or detection errors will have a significant influence on the IOU calculation of small objects, although this may not be an issue for large objects. Thus, it is reasonable to take the mAP score with IOU 0.25 into consideration for small objects.

%In Fig.\ref{Golf Ball Trace},labeled ground truth is represented by green bounding box, while the tracked result bounding box is in red. The size of the golf ball in the first frame is around $27\times27$. It was labeled 27 by 26. Suppose the prediction from the detectors are exact rig. The label error could possibly lower the iou by around 4 percent. When the golf ball gets further away, the size of the golf ball in the image decreased to $6\times6$. The golf ball was labled 5 by 6, which could cause an iou drop of 17 percent. Above only shows the error caused by 1 pixel difference on either width or height. What if the golf ball was labeled off by 1 pixel on both width and height. The possibly iou loss could be 8 percent in the first frame. For the last frame, the iou decrease would be 31 percent. These are the possible error caused by labeling off for 1 pixel. Labeling not only affects detection precision, but also affects model precision. Above all labeling error is significant for such tiny object. It is reasonable to take the mAP score with iou 0.25 into consideration to compensate for the error.  

\section{Conclusion}
The paper has proposed and implemented a practical approach for real-time golf ball detection and tracking. The proposed solution is based on object detection and the discrete Kalman filter. For object detection, we have implemented the convolutional neural networks, including YOLOv3,  YOLOv3 tiny, YOLOv4, Faster R-CNN, SSD, and RefineDet. For tracking, we have tested 5 models and Faster R-CNN yields the best performance on success rate and precision rate, while YOLOv3 tiny achieves the fastest inference rate with competing accuracy. The other 3 models yield acceptable precision and frame rate. Extensive experimental evaluations have demonstrated the effectiveness of the proposed approach, while most classical methods may fail for the detection and tracking of such a small object. The performance could be further improved if a larger and more representative training dataset is available.

\balance
{\small
	\bibliographystyle{IEEEtranS}
	\bibliography{ref}

% Generated by IEEEtranS.bst, version: 1.14 (2015/08/26)
\begin{thebibliography}{10}
\providecommand{\url}[1]{#1}
\csname url@samestyle\endcsname
\providecommand{\newblock}{\relax}
\providecommand{\bibinfo}[2]{#2}
\providecommand{\BIBentrySTDinterwordspacing}{\spaceskip=0pt\relax}
\providecommand{\BIBentryALTinterwordstretchfactor}{4}
\providecommand{\BIBentryALTinterwordspacing}{\spaceskip=\fontdimen2\font plus
\BIBentryALTinterwordstretchfactor\fontdimen3\font minus
  \fontdimen4\font\relax}
\providecommand{\BIBforeignlanguage}[2]{{%
\expandafter\ifx\csname l@#1\endcsname\relax
\typeout{** WARNING: IEEEtranS.bst: No hyphenation pattern has been}%
\typeout{** loaded for the language `#1'. Using the pattern for}%
\typeout{** the default language instead.}%
\else
\language=\csname l@#1\endcsname
\fi
#2}}
\providecommand{\BIBdecl}{\relax}
\BIBdecl

\bibitem{bochkovskiy2020yolov4}
A.~Bochkovskiy, C.-Y. Wang, and H.-Y.~M. Liao, ``Yolov4: Optimal speed and
  accuracy of object detection,'' \emph{arXiv preprint arXiv:2004.10934}, 2020.

\bibitem{DBLP:conf/cvpr/DalalT05}
\BIBentryALTinterwordspacing
N.~Dalal and B.~Triggs, ``Histograms of oriented gradients for human
  detection,'' in \emph{2005 {IEEE} Computer Society Conference on Computer
  Vision and Pattern Recognition {(CVPR} 2005), 20-26 June 2005, San Diego, CA,
  {USA}}.\hskip 1em plus 0.5em minus 0.4em\relax {IEEE} Computer Society, 2005,
  pp. 886--893. [Online]. Available:
  \url{https://doi.org/10.1109/CVPR.2005.177}
\BIBentrySTDinterwordspacing

\bibitem{10.1109/CVPR.2014.143}
\BIBentryALTinterwordspacing
M.~Danelljan, F.~S. Khan, M.~Felsberg, and J.~v.~d. Weijer, ``Adaptive color
  attributes for real-time visual tracking,'' in \emph{Proceedings of the 2014
  IEEE Conference on Computer Vision and Pattern Recognition}, ser. CVPR
  ’14.\hskip 1em plus 0.5em minus 0.4em\relax USA: IEEE Computer Society,
  2014, p. 1090–1097. [Online]. Available:
  \url{https://doi.org/10.1109/CVPR.2014.143}
\BIBentrySTDinterwordspacing

\bibitem{duan2019centernet}
K.~Duan, S.~Bai, L.~Xie, H.~Qi, Q.~Huang, and Q.~Tian, ``Centernet: Keypoint
  triplets for object detection,'' in \emph{Proceedings of the IEEE
  International Conference on Computer Vision}, 2019, pp. 6569--6578.

\bibitem{everingham2015pascal}
M.~Everingham, S.~A. Eslami, L.~Van~Gool, C.~K. Williams, J.~Winn, and
  A.~Zisserman, ``The pascal visual object classes challenge: A
  retrospective,'' \emph{International journal of computer vision}, vol. 111,
  no.~1, pp. 98--136, 2015.

\bibitem{everingham2010pascal}
M.~Everingham, L.~Van~Gool, C.~K. Williams, J.~Winn, and A.~Zisserman, ``The
  pascal visual object classes (voc) challenge,'' \emph{International journal
  of computer vision}, vol.~88, no.~2, pp. 303--338, 2010.

\bibitem{DBLP:journals/corr/Girshick15}
\BIBentryALTinterwordspacing
R.~B. Girshick, ``Fast {R-CNN},'' \emph{CoRR}, vol. abs/1504.08083, 2015.
  [Online]. Available: \url{http://arxiv.org/abs/1504.08083}
\BIBentrySTDinterwordspacing

\bibitem{he2015spatial}
K.~He, X.~Zhang, S.~Ren, and J.~Sun, ``Spatial pyramid pooling in deep
  convolutional networks for visual recognition,'' \emph{IEEE transactions on
  pattern analysis and machine intelligence}, vol.~37, no.~9, pp. 1904--1916,
  2015.

\bibitem{henriques2014high}
J.~F. Henriques, R.~Caseiro, P.~Martins, and J.~Batista, ``High-speed tracking
  with kernelized correlation filters,'' \emph{IEEE transactions on pattern
  analysis and machine intelligence}, vol.~37, no.~3, pp. 583--596, 2014.

\bibitem{Kim2011AutomaticHF}
J.-S. Kim and M.-G. Kim, ``Automatic high-speed flying ball detection from
  multi-exposure images under varying light conditions,'' in \emph{VRCAI},
  2011.

\bibitem{law2018cornernet}
H.~Law and J.~Deng, ``Cornernet: Detecting objects as paired keypoints,'' in
  \emph{Proceedings of the European Conference on Computer Vision (ECCV)},
  2018, pp. 734--750.

\bibitem{li2019siamrpn++}
B.~Li, W.~Wu, Q.~Wang, F.~Zhang, J.~Xing, and J.~Yan, ``Siamrpn++: Evolution of
  siamese visual tracking with very deep networks,'' in \emph{Proceedings of
  the IEEE Conference on Computer Vision and Pattern Recognition}, 2019, pp.
  4282--4291.

\bibitem{li20202}
K.~Li, W.~Ma, U.~Sajid, Y.~Wu, and G.~Wang, ``Object detection with
  convolutional neural networks,'' \emph{Deep Learning in Computer Vision:
  Principles and Applications}, vol.~30, no.~31, p.~41, 2020.

\bibitem{lin2017feature}
T.-Y. Lin, P.~Doll{\'a}r, R.~Girshick, K.~He, B.~Hariharan, and S.~Belongie,
  ``Feature pyramid networks for object detection,'' in \emph{Proceedings of
  the IEEE conference on computer vision and pattern recognition}, 2017, pp.
  2117--2125.

\bibitem{lin2017focal}
T.-Y. Lin, P.~Goyal, R.~Girshick, K.~He, and P.~Doll{\'a}r, ``Focal loss for
  dense object detection,'' in \emph{Proceedings of the IEEE international
  conference on computer vision}, 2017, pp. 2980--2988.

\bibitem{lin2014microsoft}
T.-Y. Lin, M.~Maire, S.~Belongie, J.~Hays, P.~Perona, D.~Ramanan,
  P.~Doll{\'a}r, and C.~L. Zitnick, ``Microsoft coco: Common objects in
  context,'' in \emph{European conference on computer vision}.\hskip 1em plus
  0.5em minus 0.4em\relax Springer, 2014, pp. 740--755.

\bibitem{liu2018path}
S.~Liu, L.~Qi, H.~Qin, J.~Shi, and J.~Jia, ``Path aggregation network for
  instance segmentation,'' in \emph{Proceedings of the IEEE conference on
  computer vision and pattern recognition}, 2018, pp. 8759--8768.

\bibitem{DBLP:journals/corr/LiuAESR15}
\BIBentryALTinterwordspacing
W.~Liu, D.~Anguelov, D.~Erhan, C.~Szegedy, S.~E. Reed, C.~Fu, and A.~C. Berg,
  ``{SSD:} single shot multibox detector,'' \emph{CoRR}, vol. abs/1512.02325,
  2015. [Online]. Available: \url{http://arxiv.org/abs/1512.02325}
\BIBentrySTDinterwordspacing

\bibitem{8296246}
C.~{Lu}, S.~{Xia}, W.~{Huang}, M.~{Shao}, and Y.~{Fu}, ``Circle detection by
  arc-support line segments,'' in \emph{2017 IEEE International Conference on
  Image Processing (ICIP)}, Sep. 2017, pp. 76--80.

\bibitem{7279260}
C.~{Lyu}, Y.~{Liu}, B.~{Li}, and H.~{Chen}, ``Multi-feature based high-speed
  ball shape target tracking,'' in \emph{2015 IEEE International Conference on
  Information and Automation}, 2015, pp. 67--72.

\bibitem{ma2020mdfn}
W.~Ma, Y.~Wu, F.~Cen, and G.~Wang, ``Mdfn: Multi-scale deep feature learning
  network for object detection,'' \emph{Pattern Recognition}, vol. 100, p.
  107149, 2020.

\bibitem{DBLP:journals/corr/RedmonF16}
\BIBentryALTinterwordspacing
J.~Redmon and A.~Farhadi, ``{YOLO9000:} better, faster, stronger,''
  \emph{CoRR}, vol. abs/1612.08242, 2016. [Online]. Available:
  \url{http://arxiv.org/abs/1612.08242}
\BIBentrySTDinterwordspacing

\bibitem{Redmon2018YOLOv3AI}
------, ``Yolov3: An incremental improvement,'' \emph{CoRR}, vol.
  abs/1804.02767, 2018.

\bibitem{ren2015faster}
S.~Ren, K.~He, R.~Girshick, and J.~Sun, ``Faster r-cnn: Towards real-time
  object detection with region proposal networks,'' 2015.

\bibitem{simonyan2014very}
K.~Simonyan and A.~Zisserman, ``Very deep convolutional networks for
  large-scale image recognition,'' \emph{arXiv preprint arXiv:1409.1556}, 2014.

\bibitem{7173015}
Y.~{Sui}, S.~{Zhang}, and L.~{Zhang}, ``Robust visual tracking via
  sparsity-induced subspace learning,'' \emph{IEEE Transactions on Image
  Processing}, vol.~24, no.~12, pp. 4686--4700, Dec 2015.

\bibitem{sui2016tracking}
Y.~Sui, G.~Wang, Y.~Tang, and L.~Zhang, ``Tracking completion,'' in
  \emph{European Conference on Computer Vision}.\hskip 1em plus 0.5em minus
  0.4em\relax Springer, 2016, pp. 194--209.

\bibitem{sui2018joint}
Y.~Sui, G.~Wang, and L.~Zhang, ``Joint correlation filtering for visual
  tracking,'' \emph{IEEE Transactions on Circuits and Systems for Video
  Technology}, 2018.

\bibitem{sui2019exploiting}
Y.~Sui, Z.~Zhang, G.~Wang, Y.~Tang, and L.~Zhang, ``Exploiting the anisotropy
  of correlation filter learning for visual tracking,'' \emph{International
  Journal of Computer Vision}, vol. 127, no.~8, pp. 1084--1105, 2019.

\bibitem{tian2019fcos}
Z.~Tian, C.~Shen, H.~Chen, and T.~He, ``Fcos: Fully convolutional one-stage
  object detection,'' in \emph{Proceedings of the IEEE International Conference
  on Computer Vision}, 2019, pp. 9627--9636.

\bibitem{uijlings2013selective}
J.~R. Uijlings, K.~E. Van De~Sande, T.~Gevers, and A.~W. Smeulders, ``Selective
  search for object recognition,'' \emph{International journal of computer
  vision}, vol. 104, no.~2, pp. 154--171, 2013.

\bibitem{article}
A.~Umek, Y.~Zhang, S.~Tomazic, and A.~Kos, ``Suitability of strain gage sensors
  for integration into smart sport equipment: A golf club example,''
  \emph{Sensors}, vol.~17, p. 916, 04 2017.

\bibitem{wang2020cspnet}
C.-Y. Wang, H.-Y. Mark~Liao, Y.-H. Wu, P.-Y. Chen, J.-W. Hsieh, and I.-H. Yeh,
  ``Cspnet: A new backbone that can enhance learning capability of cnn,'' in
  \emph{Proceedings of the IEEE/CVF Conference on Computer Vision and Pattern
  Recognition Workshops}, 2020, pp. 390--391.

\bibitem{8474387}
S.~{Wang}, Y.~{Xu}, Y.~{Zheng}, M.~{Zhu}, H.~{Yao}, and Z.~{Xiao}, ``Tracking a
  golf ball with high-speed stereo vision system,'' \emph{IEEE Transactions on
  Instrumentation and Measurement}, vol.~68, no.~8, pp. 2742--2754, Aug 2019.

\bibitem{welch1995introduction}
G.~Welch, G.~Bishop \emph{et~al.}, ``An introduction to the kalman filter,''
  1995.

\bibitem{CV}
A.~Woodward and P.~Delmas, ``Computer vision for low cost 3-d golf ball and
  club tracking,'' 3 2015.

\bibitem{Wu_Vision2017}
Y.~Wu, Y.~Sui, and G.~Wang, ``Vision-based real-time aerial object localization
  and tracking for uav sensing system,'' \emph{IEEE Access}, vol.~5, pp.
  23\,969--23\,978, 2017.

\bibitem{WuLimYang13}
Y.~Wu, J.~Lim, and M.-H. Yang, ``Online object tracking: A benchmark,'' in
  \emph{IEEE Conference on Computer Vision and Pattern Recognition (CVPR)},
  2013.

\bibitem{zhang2020bridging}
S.~Zhang, C.~Chi, Y.~Yao, Z.~Lei, and S.~Z. Li, ``Bridging the gap between
  anchor-based and anchor-free detection via adaptive training sample
  selection,'' in \emph{Proceedings of the IEEE/CVF Conference on Computer
  Vision and Pattern Recognition}, 2020, pp. 9759--9768.

\bibitem{zhang2018single}
S.~Zhang, L.~Wen, X.~Bian, Z.~Lei, and S.~Z. Li, ``Single-shot refinement
  neural network for object detection,'' in \emph{Proceedings of the IEEE
  conference on computer vision and pattern recognition}, 2018, pp. 4203--4212.

\bibitem{9283312}
X.~{Zhang}, T.~{Zhang}, Y.~{Yang}, Z.~{Wang}, and G.~{Wang}, ``Real-time golf
  ball detection and tracking based on convolutional neural networks,'' in
  \emph{2020 IEEE International Conference on Systems, Man, and Cybernetics
  (SMC)}, 2020, pp. 2808--2813.

\bibitem{zhou2019bottom}
X.~Zhou, J.~Zhuo, and P.~Krahenbuhl, ``Bottom-up object detection by grouping
  extreme and center points,'' in \emph{Proceedings of the IEEE Conference on
  Computer Vision and Pattern Recognition}, 2019, pp. 850--859.

\bibitem{zhu2019feature}
C.~Zhu, Y.~He, and M.~Savvides, ``Feature selective anchor-free module for
  single-shot object detection,'' in \emph{Proceedings of the IEEE Conference
  on Computer Vision and Pattern Recognition}, 2019, pp. 840--849.

\end{thebibliography}
}

\balance

\end{document}